\documentclass[10pt,twocolumn,letterpaper]{article}

\usepackage{geometry}
\usepackage{booktabs} 
\usepackage{bbm}
\usepackage{mathtools}
\usepackage{nccmath}
\usepackage{setspace}

\usepackage{caption}
\usepackage{subcaption}

\usepackage[linesnumbered,ruled,vlined]{algorithm2e}

\SetKwInput{KwInput}{Input}                % Set the Input
\SetKwInput{KwOutput}{Output}              % set the Output

\SetCommentSty{mycommfont}

\SetAlCapSty{algcapsty}

\usepackage[T1]{fontenc}
\usepackage{wrapfig,lipsum,booktabs}

\usepackage{soul}
\usepackage{dsfont}
\usepackage{enumerate}
\usepackage{enumitem}

\usepackage{amsmath}
\usepackage{amsfonts}
\usepackage{bbm}
\usepackage{dsfont}
\usepackage[Symbol]{upgreek}
\usepackage{lscape}
\usepackage{caption}
\usepackage{balance}
\usepackage{xspace}
\usepackage{float}
\usepackage{kotex}

\usepackage{wasysym}
\usepackage[table,xcdraw,dvipsnames]{xcolor}
\usepackage{multirow}
\usepackage{array, boldline, rotating}

\usepackage{amssymb}% http://ctan.org/pkg/amssymb
\usepackage{pifont}% http://ctan.org/pkg/pifont
\newcommand{\omark}{\textbf{$\mathcal{O}$}\xspace}%
\newcommand{\xmark}{\ding{55}\xspace}%

\newcommand{\mc}[1]{\mathcal{#1}}

\renewcommand*\eqref[1]{(\ref{#1})}

\newcommand{\eg}{\emph{e.g.,~}}
\newcommand{\ie}{\emph{i.e.,~}}

\newcommand{\myparagraph}[1]{\vspace{0.07cm}\noindent\textbf{#1}~}

\def\code#1{\texttt{#1}}

\newcommand\scalemath[2]{\scalebox{#1}{\mbox{\ensuremath{\displaystyle #2}}}}

\definecolor{LightCyan}{rgb}{0.88,1,1}
\definecolor{Blue}{rgb}{0, 0.5, 1}
\definecolor{Green}{rgb}{0.0, 0.8, 0.0 }
\definecolor{Red}{rgb}{0.95, 0.55, 0.6}
\definecolor{Skyblue}{rgb}{0.6, 0.6, 0.95 }

\NewDocumentEnvironment{suptitle}{ +b }{
    \twocolumn[{#1}]%
}{}

\NewDocumentCommand{\supptitle}{s}{
\begin{suptitle}
        \centering
        \rule{\textwidth}{0.03cm}\\[0.1cm]
        -Supplementary Material-\\[0.2cm]
        {\Large 
            \textbf{\mytitle }
        }\\%[0.40cm]
        \rule{\textwidth}{0.03cm}\\[0.2cm]
\end{suptitle}}
\usepackage{amsmath,amsfonts,bm}

\def\eqref#1{equation~\ref{#1}}
\def\1{\bm{1}}

\DeclareMathAlphabet{\mathsfit}{\encodingdefault}{\sfdefault}{m}{sl}
\SetMathAlphabet{\mathsfit}{bold}{\encodingdefault}{\sfdefault}{bx}{n}

\DeclareMathOperator*{\argmax}{arg\,max}
\DeclareMathOperator*{\argmin}{arg\,min}

\newcommand{\alg}{\code{PCB}\xspace}
\newcommand{\mytitle}{Active Prompt Learning in Vision Language Models}
\newcommand\blfootnote[1]{%
  \begingroup
  \renewcommand\thefootnote{}\footnote{#1}%
  \addtocounter{footnote}{-1}%
  \endgroup
}
\usepackage[pagenumbers]{cvpr} % To force page numbers, e.g. for an arXiv version
\usepackage[dvipsnames]{xcolor}

\definecolor{cvprblue}{rgb}{0.21,0.49,0.74}
\usepackage[pagebackref,breaklinks,colorlinks,citecolor=cvprblue]{hyperref}
\usepackage{graphicx}
\usepackage{amsmath}
\usepackage{amssymb}
\usepackage{booktabs}
\usepackage{algpseudocode}
\usepackage{color}
\usepackage{xcolor}
\usepackage{colortbl}
\usepackage[export]{adjustbox}

\def\paperID{10038} % *** Enter the Paper ID here
\def\confName{CVPR}
\def\confYear{2024}

\usepackage[capitalize]{cleveref}
\crefname{section}{Sec.}{Secs.}
\Crefname{section}{Section}{Sections}
\Crefname{table}{Table}{Tables}
\crefname{table}{Tab.}{Tabs.}

\title{\mytitle}
\author{ Jihwan Bang$^{1}$\hspace{2em}Sumyeong Ahn$^{2}$\hspace{2em}Jae-Gil Lee$^{1 *}$\\
{KAIST$^1$\hspace{4em}Michigan State University$^2$}\\
{\tt\small {\{jihwan.bang, jaegil\}@kaist.ac.kr, sumyeong@msu.edu}}
}

\begin{document}

\maketitle

\setlength{\textfloatsep}{5pt}
\blfootnote{\hspace{-1.8em}$^*$ indicates corresponding author.}
%%%%%%%%% ABSTRACT
\begin{abstract}
% Thanks to the huge progress of the pre-trained models, .....
% With the substantial advancements in pre-trained vision language models (VLMs), these models now possess the capability to perform zero-shot learning tasks. Yet, when it comes to performance, these zero-shot models still have a gap with models tailored for specific tasks. Some solutions to improve their performance are proposed by fine-tuning them with a small number of examples, but these solutions assume that we can get an equal number of examples for training from each class - which is quite difficult with unlabeled data, but understudied. 

Pre-trained Vision Language Models (VLMs) have demonstrated notable progress in various zero-shot tasks, such as classification and retrieval. Despite their performance, because improving performance on new tasks requires task-specific knowledge, their adaptation is essential. While labels are needed for the adaptation, acquiring them is typically expensive. To overcome this challenge, active learning, a method of achieving a high performance by obtaining labels for a small number of samples from experts, has been studied. Active learning primarily focuses on selecting unlabeled samples for labeling and leveraging them to train models. In this study, we pose the question, ``how can the pre-trained VLMs be adapted under the active learning framework?'' In response to this inquiry, we observe that (1) simply applying a conventional active learning framework to pre-trained VLMs even may degrade performance compared to random selection because of the class imbalance in labeling candidates, and (2) the knowledge of VLMs can provide hints for achieving the balance before labeling. Based on these observations, we devise a novel active learning framework for VLMs, denoted as \alg. To assess the effectiveness of our approach, we conduct experiments on seven different real-world datasets, and the results demonstrate that \alg surpasses conventional active learning and random sampling methods. Code will be available in \href{https://github.com/kaist-dmlab/pcb}{https://github.com/kaist-dmlab/pcb}.

\end{abstract}

%%%%%%%%% BODY TEXT
\section{Introduction}
\label{sec:intro}
In the past, as emerging research in deep neural networks (DNNs) progressed, there was a substantial focus on studying specific types of datasets, including image/video (vision)~\cite{dosovitskiy2020image, arnab2021vivit, he2016deep}, natural language~\cite{brown2020language, touvron2023llama, touvron2023llama2}, graph~\cite{yun2019graph}, table~\cite{yang2022tableformer}, and more. However, recent research has raised the question: \emph{``can we develop DNNs capable of understanding multiple types of datasets interactively?''} Among various candidates for multi-modality models, vision language models (VLMs)~\cite{radford2021learning, li2021align, yao2021filip, li2022blip, li2023blip} have garnered significant attention due to not only to their wide domain knowledge but also to their superior performance on various tasks.

Most of VLMs, for instance CLIP~\cite{radford2021learning}, comprises two encoders: image and text encoders. They have consistently shown impressive zero-shot performance across a wide range of tasks without fine-tuning. For example, CLIP is well-known for its remarkable zero-shot classification performance on various benchmarks, even if the model has not encountered the datasets previously. Despite these notable zero-shot performances, many researchers are focusing on developing adaptation methods for new target tasks because of necessity to make the model aware of the target tasks. Since updating all parameters can be computationally expensive, a key research focus lies in reducing the adaptation computing cost~\cite{khattak2023maple, zhou2022conditional, zhou2022learning}. For example, CoOp~\cite{zhou2022conditional} takes the approach of freezing both encoders and only allowing a small number of trainable parameters (with a size ranging from $4$ to $16$) to serve as prompts. This strategy has demonstrated substantial improvements in classification performance with only a small number of trainable parameters and a limited amount of data for each class.

% These successes primarily results from the capability of language models to easily establish correlations between words~\cite{pratt2023does, menon2023visual}.

% \begin{figure*}
%     \centering
%     \includegraphics[width=0.95\textwidth]{figures/AL_VLM_figure_nov13.pdf}
%     \caption{\textbf{The main difference between \textcolor{Blue}{(a) conventional active learning} and \textcolor{Green}{(b) active prompt learning in VLMs}}. Conventional active learning only utilizes the embeddings from the one model, which is fully trained by labeled data $\mathcal{D}_l$. On the other hand, active prompt learning in VLMs utilizes embeddings of both image encoder and text encoder with descriptions of each class, and only learnable contexts are trained by labeled data $\mathcal{D}_l$. Unlike conventional active learning, it needs only a few samples to train, but it is hard to get class balanced samples from unlabeled data $\mathcal{D}_u$. For example, it can be occurred that there is no car image in $\mathcal{D}_l$.}
%     % Please note that \emoji{fire} and \emoji{snowflake} indicates the trainable and frozen parameters, respectively. }
%     \label{fig:intro}
% \end{figure*}

\begin{figure*}
    \centering
    \includegraphics[width=0.95\textwidth]{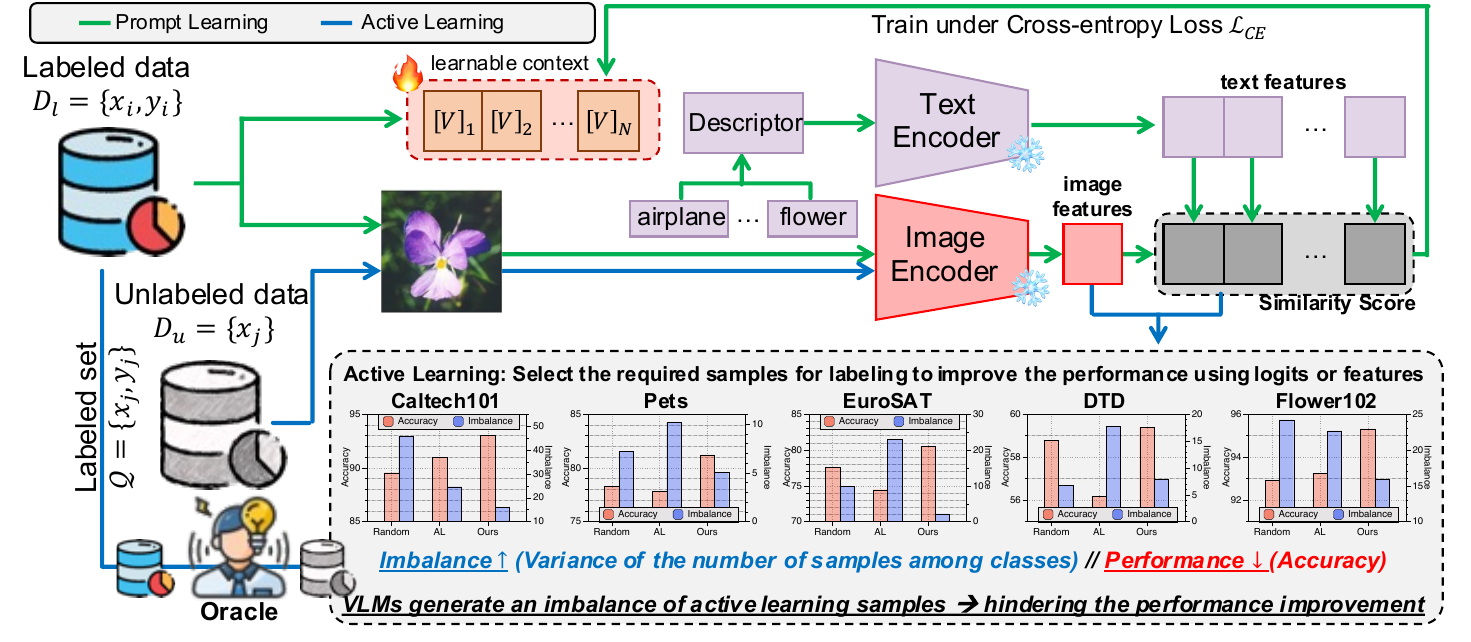}
    \vspace*{-0.3cm}
    \caption{\textbf{Key motivation and complete process behind active prompt learning.} When we emply a traditional active learning framework for adapting prompt learning to a new target task, the active learning sampler incurs a significant imbalance (indicated by \textcolor{Red}{red bars}). Thus, this imbalance results in an inability to enhance the ultimate performance (as indicated by \textcolor{Skyblue}{blue bars}). In this paper, we introduce a novel algorithm named \alg that rectifies this imbalance by harnessing the knowledge of VLMs, enabling effective utilization of the oracle.}
    \label{fig:intro}
    \vspace*{-0.4cm}
\end{figure*}

Even though we can reduce the adpation cost, the barrier of high labeling costs still persists. To mitigate this inefficiency, there have been extensive studies in an area of active learning~\cite{settles2009active, ren2021survey}. The central objective of active learning is to select samples for labeling so that the model performance is significantly improved, and making a noticebale gap compared to random samples of the same quantity. These active learning methods can be roughly divided into two categories: (1) uncertainty-based sampling~\cite{gal2017deep, houlsby2011bayesian, kirsch2019batchbald, rakesh2021efficacy, holub2008entropy} and (2) diversity-based sampling~\cite{sener2018active, parvaneh2022active} which leverages feature embeddings from the image encoder. In a hybrid perspective, BADGE~\cite{ash2019deep} was introduced by combining uncertainty and diversity through the use of $k$-means++ clustering within the gradient embedding space.

% \begin{figure}[t] 
%      \centering
%      \begin{subfigure}[b]{0.235\textwidth}
%          \centering
%          % \includegraphics[width=\textwidth]{figures/fig1_flowers.pdf}
%          \includegraphics[width=\textwidth]{figures/fig1_flower_new.pdf}
%          \caption{Flowers102}
%          \label{fig:fig1_flowers}
%      \end{subfigure}
%      \hfill
%      \begin{subfigure}[b]{0.235\textwidth}
%          \centering
%          % \includegraphics[width=\textwidth]{figures/fig1_dtd.pdf}
%          \includegraphics[width=\textwidth]{figures/fig1_dtd_new.pdf}
%          \caption{DTD}
%          \label{fig:fig1_dtd}
%      \end{subfigure}
%      \vspace{-20pt}
%      \caption{Accuracy and class imbalance in active prompt learning of VLM.}
%      \vspace{-20pt}
%      % It indicates the accuracy and class imbalance of labeled datasets after final round of active learning. 
%      % The class distribution is inconsistent and consistent in (a) and (b), respectively.}
%      \label{fig:problem}
% \end{figure} 

Under these two researches, 
% our initial inquiry revolves around whether a straightforward application of active learning for VLMs can successfully yield improved classification performance. If not, what is the key mis-alignment when combining these two approaches? 
our initial inquiry pertains to the determination of whether the implementation simply combining active learning with VLMs can effectively lead to enhanced classification performance. If it does not result in such improvement, what constitutes the critical incongruity in integrating these two methodologies?
To address this question, we observe two phenomena: (1) na\"ively applying active learning to VLMs does not consistently demonstrate improvements compared to random selection-based labeling (depicted as \textcolor{Red}{red bars} in~\autoref{fig:intro}); (2) this lack of improvement comes from the imbalanced class labels misled by an active learning framework (illustrated as \textcolor{Skyblue}{blue bars} in~\autoref{fig:intro}). The imbalanced behavior of active learning algorithms is due to the imbalanced pre-trained knowledge of VLMs. We verify that pre-trained CLIP has different knowledge of each class by showing the class-wise accuracy (see \autoref{app:imbalance}). Therefore, it is imperative to investigate how VLMs can effectively collaborate with active learning frameworks, particularly given that VLMs exacerbate the issue of class imbalance.

In this study, we introduce our approach, called \alg, which is designed to address the class imbalance issue and improve the classification performance of VLMs using only a limited amount of labeled data from experts. Our contributions are summarized as follows:

\begin{itemize}
    \item This study represents the first exploration of synergistic approaches to active learning and VLMs, marking a novel contribution to the field. We establish that a straightforward combination of these two approaches does not consistently lead to an improvement, highlighting the need for enhacing active learning methods in this context.
    \item We delve into the underlying reasons for performance degradation of conventional active learning methods when combined with VLMs. Our investigation reveals that the selection of samples to be labeled by experts is imbalanced, thereby making VLMs biased.
    \item We introduce an algorithm named \alg, which harnesses the valuable pre-trained knowledge of VLMs to address the issue of class imbalance. This algorithm seamlessly integrates with conventional active learning techniques, resulting in a substantial enhancement of active prompt learning within VLMs.
    % It is briefly illustrated in~\autoref{fig:intro}. 
\end{itemize}

\section{Problem Formulation}
\label{sec:prob}
\subsection{Active Learning}
% The objective of active learning is to facilitate the learning of a multi-class classification model with $K$ classes. 
The objective of active learning is to facilitate the learning of a multi-class classification model with $K$ classes while minimizing the labeling budget.
The model undergoes an iterative training process through interactions with an oracle, who provides correct annotations. In each iteration, the model learns from an annotated dataset, denoted as $\mc{D}_{l}$$ = \{(x_i, y_i)\}_{i=1}^{L}$, where $x_i \in \mc{X}$ is the input (\eg images), and $y_i \in \{1,...,K\}$ is the corresponding label, and $L$ is the number of labeled samples. Upon sufficiently training the model with the given dataset $\mc{D}_{l}$, the active learning algorithm selects $N$ samples from the unlabeled dataset, $\mc{D}_{u}$. The oracle then provides the labels for these selected samples, and they are subsequently incorporated into the annotated dataset $\mc{D}_{l}$ for use in the training.

\subsection{Vision Language Models and Prompt Learning}
\myparagraph{Vision language models (VLMs).}
VLMs typically consist of two encoders: an image encoder and a text encoder. When presented with an image as an input, the image encoder transforms the image into an embedding vector. In the case of CLIP~\cite{radford2021learning}, one of the most representative VLMs, it employs the ResNet~\cite{he2016deep} and ViT~\cite{dosovitskiy2020image} architectures as its image encoder. On the other hand, the objective of the text encoder is to map a sentence into an embedding vector with the same dimension as the output of the image encoder. CLIP employs the Transformer~\cite{vaswani2017attention} architecture for its text encoder. The CLIP model is trained using an Image-Text constrastive loss to align the embeddings of image and text pairs, enabling it to learn meaningful associations between images and corresponding text descriptions.

Indeed, the classification process in the CLIP model relies on the similarity score between image and text pairs. Here is a summary of how the CLIP model performs classification. Given an image $x_i$, the embeddings for the image and text are formulated as 
\begin{equation*}
    e_{\mathrm{img}} = \mathrm{CLIP_{img}}(x_i), \quad e_{\mathrm{txt}}^k = \mathrm{CLIP_{txt}}(T(\mathrm{CLS}_k)).
\end{equation*}
Here, $T$ represents the text template (\eg \emph{A photo of} $\{\mathrm{CLS}_k\}$), and $\{\mathrm{CLS}_k\}$ denotes the class name of each class index $k \in \{1, 2, ..., K\}$. Using $e_{\mathrm{img}}$ and $e_{\mathrm{txt}}$, the prediction probability of each class $k$ is formulated as 
\begin{equation*}
    \mathrm{P}(y=k|x) = \frac{\mathrm{exp}(\mathrm{cos}(e_{\mathrm{img}}, e_{\mathrm{txt}}^k)/\tau)}{\sum_{i=1}^{K}\mathrm{exp}(\mathrm{cos}(e_{\mathrm{img}}, e_{\mathrm{txt}}^i)/\tau)}
\end{equation*}
where $\tau$ represents a temperature parameter, and $\mathrm{cos(\cdot, \cdot)}$ is the cosine similarity.

\myparagraph{Prompt learning (PL).}
PL is an efficient adaptation method that allows partial parts of prompts to be trainable~\cite{zhou2022learning, zhou2022conditional, khattak2023maple, ju2022prompting, lu2022prompt, shin2020autoprompt, jiang2020can, li2021prefix, zhong2021factual, lester2021power, gao2020making}. It improves performance by updating the following criteria. Suppose that the class name $\{\mathrm{CLS}\}$ is tokenized as $[\mathrm{CLS}]$, and the text template function $T$ generates tokens as follows:
\begin{equation*}
    T(\mathrm{CLS}_k) = [V]_1[V]_2\ldots[V]_M[\mathrm{CLS}_k].
\end{equation*}
Here, $[V]_i$ represent trainable tokens, and $[\mathrm{CLS}_k]$ is a fixed token for each class name $\{\mathrm{CLS}_k\}$. Note that the position of trainable tokens can be changed, \eg by placing them right after the class token; we simply notate it as the \emph{front case} for simplification. These trainable parameters are trained using the cross-entropy loss function defined as
\begin{equation*}
    \mathcal{L}_{\text{CE}}(x_i, y_i) = - \sum_{k=1}^{K} \mathbbm{1}\{y_i = k\} \log P(y=k|x_i).
\end{equation*}

\section{Method: \alg}
\label{sec:method}
% \jhnote{Alg.\ref{alg:active_learning} 에 대해서 간략히 설명.}

As depicted in~\autoref{fig:intro}, two key motivations can be derived: (1) the traditional approach to select unlabeled samples in active learning leads to an imbalance under pre-trained VLMs, and (2) achieving a balance is imperative for enhancing overall performance, but it is challenging with unlabeled data; it can rather cause deeper imbalance by using false knowledge. Building upon the insights from these findings, we recognize the significance of balancing in improving performance within the active prompt learning problem. To address this objective, we introduce a novel algorithm named \alg: \textbf{\underline{P}}seudo-\textbf{\underline{C}}lass \textbf{\underline{B}}alance for Active Prompt Learning in VLMs. 
% A concise overview of this algorithm can be found in~\autoref{alg:active_learning}. 
In the following section, we delve into a detailed explanation of the entire workflow encompassed by the proposed algorithm.

\subsection{\underline{P}seudo-\underline{C}lass \underline{B}alance for Active Prompt Learning in VLMs}

\begin{algorithm}[t]
    \DontPrintSemicolon
    \SetAlgoLined
    \SetNoFillComment
    \LinesNotNumbered 
    \caption{\code{Balance\_sampler}}
    \label{alg:balancing}
    \KwInput{Labeled dataset $\mc{D}_{l}$, Pseudo-labeled dataset $\tilde{P}$, Budget $N$ }
    \textbf{Init:} $\mc{Q} = \emptyset$ (Query set), $\tilde{\mc{D}}_{l} = \mc{D}_{l}$ (Estimated $\mc{D}_{l}$) \\
    \For {$n = 1,2,...,N$}
    {
        \small{\textcolor{Skyblue}{\# Select class $k$, the smallest \# of class samples in $\tilde{\mc{D}}_{l}$ }} \\
        $k = \argmin_{k\in\{1,...,K\}} |c_k|$ \\
        \hfill where $c_k = \{ (x_i, y_i) | y_i = k \, \text{and} \, (x_i, y_i) \in \tilde{\mc{D}}_{l}\}$ \\
        \small{\textcolor{Skyblue}{\# Select one sample pseudo-labeled as $k$ from $\tilde{P}$}} \\
        $(x_j,\tilde{y}_j) \in \tilde{P}$, where $\tilde{y}_j = k$\\
        \small{\textcolor{Skyblue}{\# Update query set $\mc{Q}$ by adding the selected sample}} \\
        $\mc{Q} = \mc{Q} \cup [(x_j)]$\\
        \small{\textcolor{Skyblue}{\# Update estimated labeled set $\tilde{\mc{D}}_{l}$}} \\
        $\tilde{\mc{D}}_{l} = \tilde{\mc{D}}_{l} \cup [(x_j, \tilde{y}_j)]$

    }
\KwOutput{Query set $\mc{Q}$}
\end{algorithm}

\myparagraph{Balance sampler.} To satisfy the class balance while selecting informative samples for improving ultimate performance, we propose the two-stage active learning method on VLMs. First, we select a subset of informative samples, $P \subset \mc{D}_{u}$ where the size of $P$ is $\gamma \times |\mc{D}_{u}|$. Here $\gamma \in [0, 1]$ is the hyperparameter that controls how progressively allow the uncertain samples to be labeled. After selecting $P$, we pseudo-label the selected samples by using VLMs' classification ability, \ie  $\tilde{P}= \{(x_i, \tilde{y}_i)\}_{i=1}^{\gamma |D_u|}$. Then, we build the query set $\mc{Q}$ by utilizing the balance sampler, as described in~\autoref{alg:balancing}, randomly selecting samples from $\tilde{P}$ so that the expected number of samples of each class is balanced.
% Note that the labeled dataset will be balanced if the pseudo-labels are correct. 

% Labeled dataset $\mc{D}_{l}$, Unlabeled dataset $\mc{D}_{u}$, Active learning round $R$, The number of query for each round $N$, Active learning sampling parameter $\gamma$, Active learning algorithm $\mc{A}$, Oracle labeler \code{Oracle}$(\cdot)$, Model $f$.

\begin{algorithm}[t]
    
    \DontPrintSemicolon
    \SetAlgoLined
    \SetNoFillComment
    \LinesNotNumbered 
    \caption{\alg}
    \label{alg:active_learning}
    \KwInput{$\mc{D}_{l}$, $\mc{D}_{u}$, $R$,  $N$, $\gamma$, $\mc{A}$, $\code{Oracle}(\cdot)$, $f$.}
    \For {$r=1, 2, ..., R$} 
        {
        \If {$r = 1$}
            {
            \small{\textcolor{Skyblue}{\# Initial query set $\mc{Q}$}} \\
            {$\mc{Q} = \texttt{random\_sample}(\mathcal{D}_u, N)$}
            }
        \Else{
            \small{\textcolor{Skyblue}{\# Select informative subset $P$}}\\
            $P = \mc{A}(\mc{D}_{u}, \gamma|\mc{D}_{u}|, f)$\\
            \small{\textcolor{Skyblue}{\# Pseudo labeling}} \\
            $\tilde{P} = \{(x_i,\tilde{y}_i) | (x_i, y_i) \in \mc{D}_{u} \, \text{and} \, \tilde{y}_i = f(x_i)\}$ \\
            \small{\textcolor{Skyblue}{\# \code{Balance\_sampler}}~(\autoref{alg:balancing})} \\
            $\mc{Q} = \code{Balance\_sampler}(\mc{D}_u, \tilde{\mc{P}}, N)$
            }
        \small{\textcolor{Skyblue}{\# Labeling by Oracle}} \\
        $\hat{\mc{Q}} = \{(x_i, y_i) | x_i \in \mc{Q} \, \text{and} \, y_i = \code{Oracle}(x_i)\}$ \\
        \small{\textcolor{Skyblue}{\# Update both sets and train prompts}~(\autoref{sec:desc_aug})}\\
        $\mathcal{D}_l$ = $\mathcal{D}_l \cup \hat{\mathcal{Q}}$, $\mathcal{D}_u$ = $\mathcal{D}_u \backslash \mathcal{Q}$ \\
        Train learnable prompts $[V]_i$ on $\mathcal{D}_l$ 
        }

\KwOutput{Final model $f$}
\end{algorithm}

\myparagraph{Proposed method.} 
Based on the balancing module, we introduce a method called \alg, which is briefly outlined in~\autoref{alg:active_learning}. This algorithm takes inputs as a labeled dataset $\mc{D}_{l}$, an unlabeled dataset $\mc{D}_{u}$, the number of active learning rounds $R$, a query budget $N$, a progressive hyperparameter $\gamma$, an active learning algorithm $\mc{A}$, an oracle labeler $\code{Oracle}$, and a VLM model $f$. 

The initial round randomly selects a query set due to insufficient information about the target dataset. From the second round, an active learning algorithm builds an informative subset $P$ and assigns pseudo-labels to its samples. The \autoref{alg:balancing} then aims to create a balanced labeled dataset. After obtaining true labels for the query set $\mc{Q}$ from an oracle, the procedure proceeds to train the parameters $[V]_i$.
% In the initial round, due to the lack of sufficient information about the target dataset, we randomly select a query set. 
% However, from the second round, the active learning algorithm to choose an informative subset $P$ and obtain the pseudo-labels of the samples in the subset $P$. Subsequently, we execute~\autoref{alg:balancing} to create an expectedly-balanced labeled dataset. After obtaining the true labels for the query set $\mc{Q}$ from the oracle, we proceed to train the trainable parameters $[V]_i$.

% To satisfy both class imbalance problem and uncertainty, we propose the two-stage active learning method. First, we filtered informative dataset $P$ ($|P| = \gamma |D_u|$) by using previous conventional active learning methods, where $\gamma \in [0, 1]$. After generating $P = \{(x_i, \tilde{y}_i)\}_{i=1}^{\gamma |D_u|}$, where $\tilde{y}_i$ indiates the pseudo label of $x_i$, we find out the class index that has the smallest number of examples in the labeled dataset $D_l$, and select the example from $P$ randomly and add it to $Q$. 
% Finally, \alg builds $Q$ by selecting examples from $P$ to balance the classes using $\tilde{y}_i$.

% \smnote{이런?
% \begin{equation*}
%     \mc{U}(P_c, U) \text{ where } P_c = \{(x_i, y_i)| y_i = c \text{ and } (x_i, y_i) \in \mc{D} \}.
% \end{equation*}
% Here, $\mc{U}(P, U)$ represents a sampling function that selects $U$ enlements from set $P$. So, ultimately select $\mc{D}_{l} = \cup_{c=1}^{K} \mc{U}(P_c, U)$.
% }

\subsection{Description Augmentation}
\label{sec:desc_aug}

In order to improve the classification performance of VLM-based models, numerous studies have explored the integration of external knowledge, as demonstrated in previous research works~\cite{menon2023visual, pratt2023does}. These studies have contributed to show how models (\eg  GPT-3~\cite{brown2020language}) can help VLMs by generating visual description for each class. For instance, the authors of~\cite{menon2023visual} introduced the following templates:

\smallskip
\setlength{\leftskip}{10pt}\noindent\texttt{Q: What are useful features for distinguishing a \{CLS\} in a photo?}

\smallskip
\noindent\texttt{A: There are several useful visual features to tell there is a \{CLS\} in a photo:}

\setlength{\leftskip}{0pt}

\smallskip
\noindent where \texttt{\{CLS\}} indicates the class name. Here, note that we can obtain $\delta_k$ descriptions for class $k$, \ie $\Delta_k = \{ \mathrm{d}_k^i\}_{i=1}^{\delta_k}$, where $\mathrm{d}_k^i$ denote $i$-th description for class $k$. See~\autoref{app:desc} for detailed prompt template.

By following the results of~\cite{menon2023visual, pratt2023does}, we adopt their prompts for training the model. In other words, we utilize the new text template function $T$ as follows:

\begin{equation*}
    \scalemath{0.96}{
    T(\mathrm{CLS}_k, i) = [V]_1...[V]_M[\mathrm{CLS}_k]\mathrm{\,\,[which]\,[is]\,\,}[\mathrm{d}_k^i],
    }
\end{equation*}

Based on this new text template function, we can use two possible prediction probabilites.

\noindent
\textbf{(1) Average Similarity (\code{AS}): }
\begin{equation*}
    \mathrm{P}(y=k|x) = \frac{1}{\delta_k} \sum_{i=1}^{\delta_k} \mathrm{P}(y=k|x, \mathrm{d}_k^i), %\label{eq:avg_score}
\end{equation*}
where 
\begin{equation*}
    \mathrm{P}(y=k|x, \mathrm{d}_k^i) = \frac{\mathrm{exp}(\mathrm{cos}(e_{\mathrm{img}}, e_{\mathrm{txt}}^{k, i})/\tau)}{\sum_{i=1}^{K}\sum_{j=1}^{\delta_k}\mathrm{exp}(\mathrm{cos}(e_{\mathrm{img}}, e_{\mathrm{txt}}^{k, i})/\tau) }.
\end{equation*}

\noindent
\textbf{(2) Average Embedding (\code{AE}): }
\begin{equation*}
    \mathrm{P}(y=k|x) = \frac{\mathrm{exp}(\mathrm{cos}(e_{\mathrm{img}}, e_{\mathrm{txt}}^k)/\tau)}{\sum_{i=1}^{K}\mathrm{exp}(\mathrm{cos}(e_{\mathrm{img}}, e_{\mathrm{txt}}^i)/\tau)},  %\label{eq:avg_emb}
\end{equation*}
where 
% $e_{\mathrm{txt}}^k  = \frac{1}{\delta_k} \sum_{i=1}^{\delta_k} e_{\mathrm{txt}}^{k, i}$.
\begin{equation}
    e_{\mathrm{txt}}^k  = \frac{1}{\delta_k} \sum_{i=1}^{\delta_k} e_{\mathrm{txt}}^{k, i}. \nonumber
\end{equation}
Note that the primary distinction between two probability scores lies in their respective averaging timeframes. \code{AS} calcuates individual embeddings and then computes the average similarity, whereas \code{AE} first averages the embeddings and then assesses the similarity.

\begin{table*}[t]

  \centering
  \begin{adjustbox}{width=0.85\linewidth}
  \begin{tabular}{@{}lcccccccc@{}}
    \toprule
     % & \multicolumn{7}{c}{\textbf{Final Accuracy ($\uparrow$)}} & \\
    \textbf{Method}                        &  \multicolumn{1}{c}{Flowers102} & \multicolumn{1}{c}{DTD} & \multicolumn{1}{c}{Oxford Pets} & \multicolumn{1}{c}{EuroSAT} &\multicolumn{1}{c}{Caltech101}  & \multicolumn{1}{c}{Stanford Cars} & \multicolumn{1}{c}{Aircraft} & \textbf{Avg Acc ($\uparrow$)}\\ \cmidrule(lr){1-1} \cmidrule(lr){2-8}\cmidrule(lr){9-9}
    CLIP (zero-shot)                    & 66.7                              & 44.5                              & 87.0                              & 49.4 
                                        & 87.9                              & 59.4                              & 21.2                              & 59.44 \\ 
    Random                              & 92.92\small{$\pm$0.61}            & 58.77\small{$\pm$1.94}            & 78.30\small{$\pm$0.74}            & 77.62\small{$\pm$1.12} 
                                        & 89.55\small{$\pm$1.00}            & 65.96\small{$\pm$0.08}            & 30.69\small{$\pm$0.30}            & 70.54 \\ \cmidrule(lr){1-9}
    Entropy~\cite{holub2008entropy}     & 94.80\small{$\pm$0.75}            & 59.18\small{$\pm$1.31}            & 76.81\small{$\pm$1.38}            & 75.46\small{$\pm$3.39} 
                                        & 91.67\small{$\pm$0.09}            & 66.68\small{$\pm$0.91}            & 25.80\small{$\pm$0.78}            & 70.06 \\
    $~~ + \mathrm{\code{AE}}$           & 96.06\small{$\pm$0.63}            & 60.80\small{$\pm$1.18}            & 78.35\small{$\pm$1.30}            & 79.97\small{$\pm$2.70}
                                        & 92.87\small{$\pm$0.20}            & 65.99\small{$\pm$0.26}            & 26.69\small{$\pm$1.34}            & 71.53 \\ 
    $~~ + \mathrm{\code{AS}}$           & 95.67\small{$\pm$1.19}            & 59.34\small{$\pm$0.81}            & 79.88\small{$\pm$1.43}            & 79.50\small{$\pm$0.60} 
                                        & 93.28\small{$\pm$0.55}            & 68.54\small{$\pm$0.09}            & 26.04\small{$\pm$1.27}            & 71.75 \\
    $~~ + \mathrm{\alg}$              & 96.16\small{$\pm$0.45}            & 59.73\small{$\pm$1.96}            & 80.44\small{$\pm$1.24}            & 80.80\small{$\pm$2.88} 
                                        & 92.41\small{$\pm$0.50}            & 67.18\small{$\pm$0.28}            & 26.78\small{$\pm$0.87}            & 71.93 \\
    $~~ + \mathrm{\alg(\code{AE})}$          & 96.33\small{$\pm$0.06}            & \underline{60.07\small{$\pm$1.69}}& 80.87\small{$\pm$0.60}            & \underline{81.72\small{$\pm$0.53}} 
                                        & 93.14\small{$\pm$0.51}            & 66.42\small{$\pm$0.86}            & 27.09\small{$\pm$0.13}            & 72.23 \\ 
    $~~ + \mathrm{\alg(\code{AS})}$          & \underline{\textbf{96.94}\small{$\pm$0.19}}   & 59.50\small{$\pm$1.99}& \underline{80.94\small{$\pm$1.05}}& 80.75\small{$\pm$1.15} 
                                        & \underline{93.48\small{$\pm$0.26}}& \underline{68.93\small{$\pm$0.86}}& \underline{27.58\small{$\pm$0.43}}& \underline{72.59} \\ \cmidrule(lr){1-9}
    Coreset~\cite{sener2018active}      & 88.65\small{$\pm$0.68}            & 50.39\small{$\pm$0.54}            & 76.70\small{$\pm$0.52}            & 68.09\small{$\pm$1.54} 
                                        & 88.78\small{$\pm$0.49}            & 61.75\small{$\pm$0.60}            & 24.32\small{$\pm$0.45}            & 65.53 \\
    $~~ + \mathrm{\code{AE}}$           & 89.06\small{$\pm$0.62}            & 51.89\small{$\pm$1.38}            & 78.08\small{$\pm$1.07}            & 68.02\small{$\pm$2.86} 
                                        & 88.99\small{$\pm$0.82}            & 60.65\small{$\pm$0.33}            & 25.88\small{$\pm$0.70}            & 66.08 \\ 
    $~~ + \mathrm{\code{AS}}$           & 89.73\small{$\pm$0.93}            & 52.76\small{$\pm$1.21}            & 78.89\small{$\pm$0.84}            & 68.07\small{$\pm$1.04} 
                                        & 90.63\small{$\pm$0.54}            & 64.15\small{$\pm$0.77}            & 26.11\small{$\pm$0.86}            & 67.19 \\
    $~~ + \mathrm{\alg}$              & 91.30\small{$\pm$0.90}            & 55.77\small{$\pm$1.33}            & 76.84\small{$\pm$1.10}            & 77.50\small{$\pm$4.64} 
                                        & 89.96\small{$\pm$0.03}            & 63.63\small{$\pm$0.27}            & 25.38\small{$\pm$0.64}            & 68.63 \\
    $~~ + \mathrm{\alg(\code{AE})}$          & 91.70\small{$\pm$0.29}            & \underline{57.09\small{$\pm$0.63}}& 78.60\small{$\pm$1.14}            & \underline{79.28\small{$\pm$0.14}} 
                                        & 90.29\small{$\pm$0.30}            & 62.08\small{$\pm$0.35}            & 26.19\small{$\pm$1.40}            & 69.31 \\ 
    $~~ + \mathrm{\alg(\code{AS})}$          & \underline{92.33\small{$\pm$0.84}}& 56.38\small{$\pm$0.73}            & \underline{79.50\small{$\pm$0.91}}& \underline{79.28\small{$\pm$1.42}} 
                                        & \underline{91.70\small{$\pm$0.48}}& \underline{65.75\small{$\pm$0.55}}& \underline{26.22\small{$\pm$0.47}}& \underline{70.17} \\
    \cmidrule(lr){1-9}
    BADGE~\cite{ash2019deep}            & 96.33\small{$\pm$0.39}            & 58.98\small{$\pm$1.30}            & 80.03\small{$\pm$1.19}            & 79.79\small{$\pm$0.94} 
                                        & 92.54\small{$\pm$0.01}            & 68.07\small{$\pm$0.61}            & 31.25\small{$\pm$0.45}            & 72.43  \\
    $~~ + \mathrm{\code{AE}}$          &  96.24\small{$\pm$0.29}            & 59.97\small{$\pm$0.71}            & 81.94\small{$\pm$0.55}            & 80.57\small{$\pm$1.40}
                                       & 92.93\small{$\pm$0.02}             & 67.10\small{$\pm$0.47}            & 31.04\small{$\pm$0.32}            & 72.83 \\ 
    $~~ + \mathrm{\code{AS}}$          &  96.44\small{$\pm$0.16}            & 61.52\small{$\pm$1.25}            & 82.33\small{$\pm$0.72}            & 81.66\small{$\pm$0.41}
                                       & 93.79\small{$\pm$0.25}             & 70.56\small{$\pm$0.31}            & 31.79\small{$\pm$0.74}            & 74.01 \\
    $~~ + \mathrm{\alg}$              & 96.12\small{$\pm$0.12}            & 60.28\small{$\pm$0.75}            & 80.22\small{$\pm$1.69}            & \underline{\textbf{81.98}\small{$\pm$0.81}}
                                        & 92.21\small{$\pm$0.92}            & 68.50\small{$\pm$0.26}            & 31.35\small{$\pm$0.21}            & 72.95 \\
    $~~ + \mathrm{\alg(\code{AE})}$          & 96.35\small{$\pm$0.27}            & 61.92\small{$\pm$1.60}            & 81.93\small{$\pm$0.88}            & 80.70\small{$\pm$3.67} 
                                        & 92.52\small{$\pm$0.32}            & 67.70\small{$\pm$0.84}            & 31.80\small{$\pm$0.08}            & 73.27  \\ 
    $~~ + \mathrm{\alg(\code{AS})}$          & \underline{96.71\small{$\pm$0.29}}& \underline{\textbf{62.33}\small{$\pm$1.06}} & \underline{\textbf{83.16}\small{$\pm$0.18}}   & 81.50\small{$\pm$1.11} 
                                        & \underline{\textbf{93.85}\small{$\pm$0.37}}& \underline{\textbf{70.70}\small{$\pm$0.79}} & \underline{\textbf{32.27}\small{$\pm$0.66}}   & \underline{\textbf{74.36}} \\ \cmidrule(lr){1-9}
    Full data                           & 97.9                              & 74.7                              & 89.3                              & 94.5 
                                        & 94.4                              & 80.8                              & 43.4                              & 82.14  \\ 
    \bottomrule
    \end{tabular}

    \end{adjustbox}
    \vspace*{-0.3cm}
  \caption{\textbf{Final accuracy on seven downstream tasks with ViT-B/32 image encoder.} Final Accuracy is the the accuracy after eight rounds, and Avg Acc is the average of final accuracies of seven datasets. \code{AS} and \code{AE} are the average score and the average embedding, respectively, as described in~\autoref{sec:desc_aug}. Also, CLIP\,(zero-shot) is the accuracy of each task from pretrained CLIP as reported in~\cite{radford2021learning}, and Full data is the accuracy when exploiting the whole dataset while prompt learning.
  Please note that the bold type and underline type represent the best performance overall and within the same active learning, respectively. For large datasets, see \autoref{app:large_dataset}.
  } 
  \vspace*{-0.5cm}
  \label{tab:mainresults}
\end{table*}

% \begin{figure*}
%      \centering
%      \includegraphics[width=0.6\textwidth]{figures/round_legend.pdf}\\
%      \begin{subfigure}[b]{0.32\textwidth}
%          \centering
%          % \includegraphics[width=\textwidth]{figures/oxford_pets.pdf}
%          \includegraphics[width=\textwidth]{figures/flower102_new.pdf}
%          \caption{Flowers102}
%          \label{fig:flower_round}
%      \end{subfigure}
%      \hfill
%      \begin{subfigure}[b]{0.32\textwidth}
%          \centering
%          % \includegraphics[width=\textwidth]{figures/dtd.pdf}
%          \includegraphics[width=\textwidth]{figures/dtd_new.pdf}
%          \caption{DTD}
%          \label{fig:dtd_round}
%      \end{subfigure}
%      \hfill
%      \begin{subfigure}[b]{0.32\textwidth}
%          \centering
%          % \includegraphics[width=\textwidth]{figures/flower102.pdf}
%          \includegraphics[width=\textwidth]{figures/oxford_pets_new.pdf}
%          \caption{Oxford Pets}
%          \label{fig:pets_round}
%      \end{subfigure}

%      \vspace*{-0.3cm}
%         \caption{\textbf{Accuracy on various downstream tasks with ViT-B/32 image encoder for each round.} For the sake of space, we only show the graphs of three datasets (others in Appendix.A). As round goes on, the accuracy increases due to increasing the number of labeled dataset $\mathcal{D}_l$. For all rounds, BADGE+\alg(AS) mostly outperforms the others, but Entropy+\alg(AS) sometimes has the best performance.}
%         \vspace*{-0.2cm}
%         \label{fig:round}
% \end{figure*}

\section{Experiment}
\label{sec:exp}

\subsection{Implementation Details} 
\myparagraph{Datasets.} For image classification in downstream tasks, we select seven openly available image classification datasets that have been previously utilized in the CLIP model~\cite{radford2021learning}, specifically EuroSAT~\cite{helber2019eurosat}, Oxford Pets~\cite{parkhi2012cats}, DTD~\cite{cimpoi2014describing}, Caltech101~\cite{fei2004learning}, Flowers102~\cite{nilsback2008automated}, StanfordCars~\cite{krause20133d}, and FGVC-Aircraft~\cite{maji2013fine}. These benchmarks span diverse categories, encompassing classification tasks involving common objects, scenes, patterns, and fine-grained categories. For more details regarding datasets, see~\autoref{app:exp}.
% Moreover, it addresses specialized tasks such as the recognition of textures and satellite imagery.

\myparagraph{Training details.} 
% Our experiment setup of active learning consists of eight rounds and each round should select the data of which size is same as the number of classes of datasets. We utilize ViT-B/32 as the image encoder's backbone, and the size of CoOp's context vectors(\ie $M$) is 16 and they are randomly initialized with zero-mean Gaussian distribution with standard deviation 0.02. While training the model for all the rounds, we utilize the SGD optimizer with learning rate 0.002, which is decayed by the cosine annealing scheduler. The epoch is set to 200. 
Our active learning setup consists of eight rounds (\ie $R$=8), and in each round, we select a subset whose size is the number of classes, \ie $N$=$K$ . To serve as the backbone for our image encoder, we adopt ViT-B/32. The size of the context vectors $M$ is set to $16$, and they are initialized using a zero-mean Gaussian distribution with a standard deviation of $0.02$.
Throughout the training process for all rounds, we employ the SGD optimizer with a learning rate $0.002$, which is decayed by the cosine annealing scheduler. We also set the maximum epoch as $200$.
All methods are implemented with PyTorch 2.0.1 and executed on a single NVIDIA A5000 GPU. 

\myparagraph{Active learning methods.} 
% Most of the active learning methods can be splitted as two parts; (1) uncertainty-based and (2) diversity-based. 
To validate the effectiveness of \alg, we select three representative active learning methods: (1) Entropy~\cite{holub2008entropy} selects the most uncertain examples with the highest entropy value from logits in the prediction;
% which is the uncertainty-based sampling method, 
(2) Coreset~\cite{sener2018active} queries the most diverse examples using embeddings from the model (\ie image encoder);
% which is the diversity-based sampling method with embedding of an image encoder,
and (3) BADGE~\cite{ash2019deep} considers both uncertainty and diversity by selecting the examples via $k$-means++ clustering in the gradient space.
% which is the hybrid sampling method considering both uncertainty and diversity. 
By adding \alg into those active learning methods, we study the synergy of active learning and \alg, and compare the results with random sampling (\ie instead of active learning) and zero-shot CLIP. Furthermore, we show the results when using the descriptions presented in Section~\ref{sec:desc_aug}. To illustrate the room for performance enhancement, we also measure the performance when prompt learning the model with the whole dataset (see ``Full data'').

\myparagraph{Metrics.} To validate the effectiveness of our method, we use the final accuracy that indicates the accuracy at the last round. As in previous analysis about imbalance, we use the variance value of the number of samples among classes. Note that all experiments are conducted three times, and all the results are reported as averages.

% Given the labeled dataset $\mathfrak{D}_l$, the class imbalance can be formulated as below: 
% variance로 적기.

% \begin{equation}
% \label{eq:balance}
%     \mathrm{imbalance} = \sqrt{\sum_{i=1}^{K} (|\mathcal{C}_i| - \frac{|\mathfrak{D}_l|}{K})^2}
% \end{equation}
% where $\mathcal{C}_k = \{(x_i, y_i) \in \mathfrak{D}_l | y_i = k \}$, and $K$ is the number of classes.

\subsection{Overall Results} 

\begin{table*}[t]
  \centering
  \begin{adjustbox}{width=0.95\linewidth}
  \begin{tabular}{@{}clcccccccc@{}}
    \toprule
    % &  & \multicolumn{7}{c}{\textbf{Final Accuracy ($\uparrow$)}} &  \\
         \textbf{Model}  &\textbf{Method}  & \multicolumn{1}{c}{Flowers102} & \multicolumn{1}{c}{DTD }& \multicolumn{1}{c}{Oxford Pets}& \multicolumn{1}{c}{EuroSAT} & \multicolumn{1}{c}{Caltech101}  & \multicolumn{1}{c}{Stanford Cars} & \multicolumn{1}{c}{Aircraft} & \textbf{Avg Acc ($\uparrow$)}\\ \cmidrule(lr){1-1} \cmidrule(lr){2-2} \cmidrule(lr){3-9} \cmidrule(lr){10-10}
    \multirow{7}{*}{RN50}              & CLIP (zero-shot)                
                                       & 65.9                             & 41.7                            & 85.4                            & 41.1                            
                                       & 82.1                             & 55.8                            & 19.3                            & 55.9 \\
    & Random                           & 92.06\small{$\pm$0.54}	         & 56.62\small{$\pm$0.97}           & 74.65\small{$\pm$0.50}          &	79.10\small{$\pm$2.31}          
                                       & 84.11\small{$\pm$0.75}	         & 61.34\small{$\pm$0.57}           & 29.15\small{$\pm$0.32}          & 68.18  \\ \cmidrule(lr){2-10}
    & BADGE~\cite{ash2019deep}         & 95.56\small{$\pm$0.54}	         & 58.35\small{$\pm$1.20}           & 75.06\small{$\pm$0.50}          &	80.94\small{$\pm$0.55}          
                                       & 89.67\small{$\pm$0.30}	         & 63.96\small{$\pm$0.53}           & 28.12\small{$\pm$1.03}          & 70.24  \\
    & $~~ + \mathrm{\alg}$           & 95.66\small{$\pm$0.28}	         & 57.41\small{$\pm$0.17}           & 76.51\small{$\pm$1.83}          &	80.06\small{$\pm$0.97}          
                                       & 89.06\small{$\pm$0.21}	         & 63.18\small{$\pm$0.77}           & 29.23\small{$\pm$0.35}          & 70.16  \\
    & $~~ + \mathrm{\alg(\code{AE})}$       & 95.72\small{$\pm$0.31}	         & \textbf{59.20}\small{$\pm$1.25}  & 76.77\small{$\pm$0.65}          & \textbf{81.96}\small{$\pm$0.60} 
                                       & 89.57\small{$\pm$0.19}          & 62.62\small{$\pm$0.26}           & 28.85\small{$\pm$1.59}          & 70.67 \\ 
    & $~~ + \mathrm{\alg(\code{AS})}$       & \textbf{96.18}\small{$\pm$0.07} & 59.14\small{$\pm$1.08}           & \textbf{80.09}\small{$\pm$0.85} & 81.60\small{$\pm$2.89}          
                                       & \textbf{90.76}\small{$\pm$0.34} & \textbf{66.20}\small{$\pm$0.69}	        & \textbf{29.61}\small{$\pm$0.78} & \textbf{71.94} \\ \cmidrule(lr){2-10}
    & Full data                        & 97.6                            & 71.6                             & 88.0                            & 93.6 
                                       & 92.8                            & 78.8                             & 42.6                            & 80.71 \\ \cmidrule(lr){1-10}
    
    \multirow{7}{*}{RN101}             & CLIP (zero-shot)                
                                       & 65.7                            & 43.9                             & 86.2                            & 33.1                            
                                       & 85.1                            & 62.3                             & 19.5                            & 56.54 \\ 
    & Random                           & 92.87\small{$\pm$0.43}	         & 58.29\small{$\pm$1.24}           & 79.08\small{$\pm$1.39}          & 77.21\small{$\pm$4.13} 
                                       & 87.55\small{$\pm$0.75}	         & 70.02\small{$\pm$0.36}           & 32.76\small{$\pm$0.29}          & 71.11 \\ \cmidrule(lr){2-10}
    & BADGE~\cite{ash2019deep}         & 96.26\small{$\pm$0.07}	         & 59.93\small{$\pm$1.25}           & 80.77\small{$\pm$1.31}          &	78.23\small{$\pm$2.22} 
                                       & 91.35\small{$\pm$0.32}	         & 71.43\small{$\pm$0.97}           & 32.56\small{$\pm$0.64}          & 72.93 \\
    & $~~ + \mathrm{\alg}$           & 95.79\small{$\pm$0.38}	         & 60.20\small{$\pm$1.89}           & 80.94\small{$\pm$0.42}          &	79.55\small{$\pm$1.37} 
                                       & 91.75\small{$\pm$0.44}	         & 71.35\small{$\pm$0.39}           & 32.62\small{$\pm$1.48}          & 73.17 \\
    & $~~ + \mathrm{\alg(\code{AE})}$       & \textbf{96.49}\small{$\pm$0.26} & \textbf{62.59}\small{$\pm$0.84}  & 83.02\small{$\pm$0.89}	      & \textbf{81.50}\small{$\pm$0.69} 
                                       & 92.51\small{$\pm$0.32}	         & 71.42\small{$\pm$0.77}	        & 32.76\small{$\pm$0.76}          & 74.33 \\ 
    & $~~ + \mathrm{\alg(\code{AS})}$       & 96.47\small{$\pm$0.18}	         & 62.17\small{$\pm$1.04}           & \textbf{83.48}\small{$\pm$2.13} & 81.14\small{$\pm$1.57}	
                                       & \textbf{92.87}\small{$\pm$0.18} & \textbf{74.04}\small{$\pm$0.39}           & \textbf{32.84}\small{$\pm$0.85} & \textbf{75.43} \\ \cmidrule(lr){2-10}
    & Full data                        & 97.8                            & 74.2                             & 91.1                            & 92.9 
                                       & 94.7                            & 83.7                             & 46.0                            & 82.91 \\ \cmidrule(lr){1-10}
    
    % \multirow{7}{*}{ViT-B/32}         & CLIP (zero-shot) & 66.7 & 44.5 & 87.0 & 49.4 & 87.9 & 59.4 & 21.2 & XX.XX \\ 
    % & Random & 92.92\small{$\pm$0.61} & 58.77\small{$\pm$1.94} & 78.30\small{$\pm$0.74} & 77.62\small{$\pm$1.12} & 89.55\small{$\pm$1.00} & 65.96\small{$\pm$0.08} & 30.69\small{$\pm$0.30} & XX.XX    \\ \cmidrule(lr){2-10}
    % &BADGE & 96.33\small{$\pm$0.39}   & 58.98\small{$\pm$1.30} & 80.03\small{$\pm$1.19} & 79.79\small{$\pm$0.94} & 92.54\small{$\pm$0.01} & 68.07\small{$\pm$0.61} & 31.25\small{$\pm$0.45} & XX.XX  \\
    % &$~~ + \mathrm{\alg}$           & 96.12\small{$\pm$0.12} & 60.28\small{$\pm$0.75} & 80.22\small{$\pm$1.69} & \textbf{81.98}\small{$\pm$0.81} & 92.21\small{$\pm$0.92} & 68.50\small{$\pm$0.26} & 31.35\small{$\pm$0.21} & XX.XX   \\
    % &$~~ + \mathrm{\alg(AE)}$       & 96.35\small{$\pm$0.27} & 61.92\small{$\pm$1.60} & 81.93\small{$\pm$0.88} & 80.70\small{$\pm$3.67} & 92.52\small{$\pm$0.32} & XX.XX\small{$\pm$0.00} & 31.80\small{$\pm$0.08} & XX.XX \\ 
    % &$~~ + \mathrm{\alg(AS)}$       & \textbf{96.71}\small{$\pm$0.29} & \textbf{62.33}\small{$\pm$1.06} & \textbf{83.16}\small{$\pm$0.18} & 81.50\small{$\pm$1.11} & \textbf{93.85}\small{$\pm$0.37} & XX.XX\small{$\pm$0.00} & \textbf{32.27}\small{$\pm$0.66} & XX.XX   \\ \cmidrule(lr){2-10}
    % & Full data                       & 97.9 & 74.7 & 89.3 & 94.5 & 94.4 & 80.8 & 43.4 & XX.XX \\  \cmidrule(lr){1-10}
    
    \multirow{7}{*}{ViT-B/16}           & CLIP (zero-shot) 
                                        & 70.4                             & 46.0                             & 88.9                               & 54.1 
                                        & 88.9                             & 65.6                             & 27.1                               & 63.0    \\ 
    & Random                            & 94.98\small{$\pm$0.06}	       & 62.63\small{$\pm$1.81}           & 84.36\small{$\pm$1.34}             & 81.14\small{$\pm$1.83} 
                                        & 90.95\small{$\pm$0.85}	       & 73.62\small{$\pm$0.30}           & 38.88\small{$\pm$0.25}             & 75.22 \\ \cmidrule(lr){2-10}
    & BADGE~\cite{ash2019deep}          & 97.97\small{$\pm$0.41}	       & 62.84\small{$\pm$2.17}           & 85.54\small{$\pm$1.30}             & 82.22\small{$\pm$1.94} 
                                        & 93.77\small{$\pm$0.51}	       & 76.55\small{$\pm$0.78}           & 39.64\small{$\pm$0.14}             & 76.93  \\
    & $~~ + \mathrm{\alg}$            & \textbf{98.32}\small{$\pm$0.21}  & 64.89\small{$\pm$1.45}           & 86.22\small{$\pm$0.71}             & 81.53\small{$\pm$3.11}	
                                        & 93.75\small{$\pm$0.28}           & 76.36\small{$\pm$0.27}           & 40.20\small{$\pm$0.30}             & 77.32   \\
    & $~~ + \mathrm{\alg(\code{AE})}$        & 98.21\small{$\pm$0.21}	       & \textbf{65.25}\small{$\pm$1.28}  & 87.23\small{$\pm$0.35}             & \textbf{84.04}\small{$\pm$2.92} 
                                        & 94.51\small{$\pm$0.29}           & 75.84\small{$\pm$0.44}           & 39.93\small{$\pm$0.21}             & 77.86 \\ 
    & $~~ + \mathrm{\alg(\code{AS})}$        & 98.19\small{$\pm$0.17}	       & 64.95\small{$\pm$1.47}           & \textbf{88.10}\small{$\pm$1.49}    & 83.85\small{$\pm$2.45}	
                                        & \textbf{95.12}\small{$\pm$0.26}  & \textbf{78.19}\small{$\pm$0.48}           &	\textbf{40.56}\small{$\pm$0.51}    & \textbf{78.42} \\ \cmidrule(lr){2-10}
    & Full data                         & 99.0                             & 77.7                             & 92.7                               & 95.1 
                                        & 95.3                             & 85.3                             & 53.6                               & 85.53    \\ 
    \bottomrule
    \end{tabular}
    \end{adjustbox}
    % \vspace*{-0.3cm}
    \vspace*{-10pt}
    \caption{\textbf{Various architectures of image encoder.} We report the performance on various types of architectures, such as ResNet-50/101 and ViT-B/16, under the BADGE active learning algorithm. The performance under Entropy and Coreset are described in~\autoref{app:additional_results}.}
  % \caption{\textbf{Ablation studies for various image encoder architectures.} The synergy of Entropy and Coreset with \alg for various image encoders is reported in Appendix.B for the space sake. \alg with description has generally the best performance for all the architectures. As the model size of the image encoder is bigger, the accuracy increases for all the methods, and gets closer to the upperbound.}
  \vspace*{-10pt}
  \label{tab:various_arch}
\end{table*}

\begin{figure}
     \centering
     \begin{subfigure}[b]{0.155\textwidth}
         \centering
         \includegraphics[width=\textwidth]{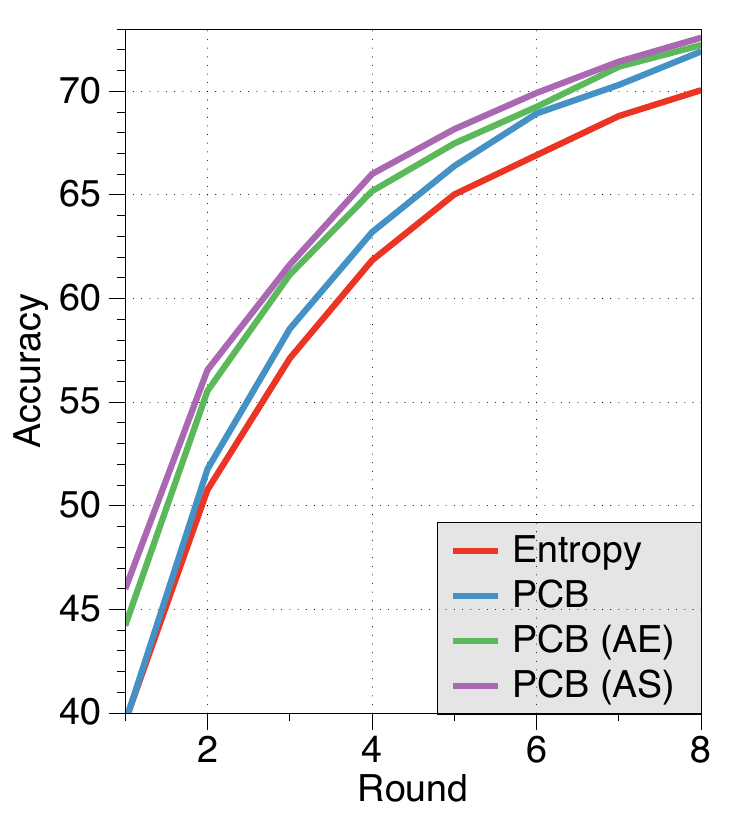}
         \vspace{-17pt}
         \caption{Entropy}
         \label{fig:entropy}
     \end{subfigure}
     \begin{subfigure}[b]{0.155\textwidth}
         \centering
         \includegraphics[width=\textwidth]{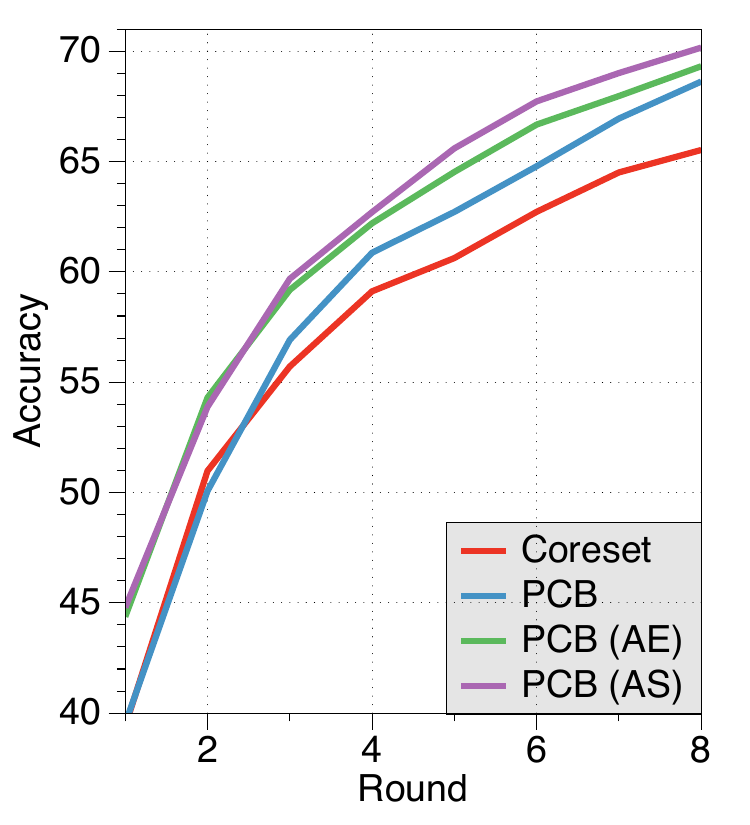}
         \vspace{-17pt}
         \caption{Coreset}
         \label{fig:coreset}
     \end{subfigure} 
     \begin{subfigure}[b]{0.155\textwidth}
         \centering
         \includegraphics[width=\textwidth]{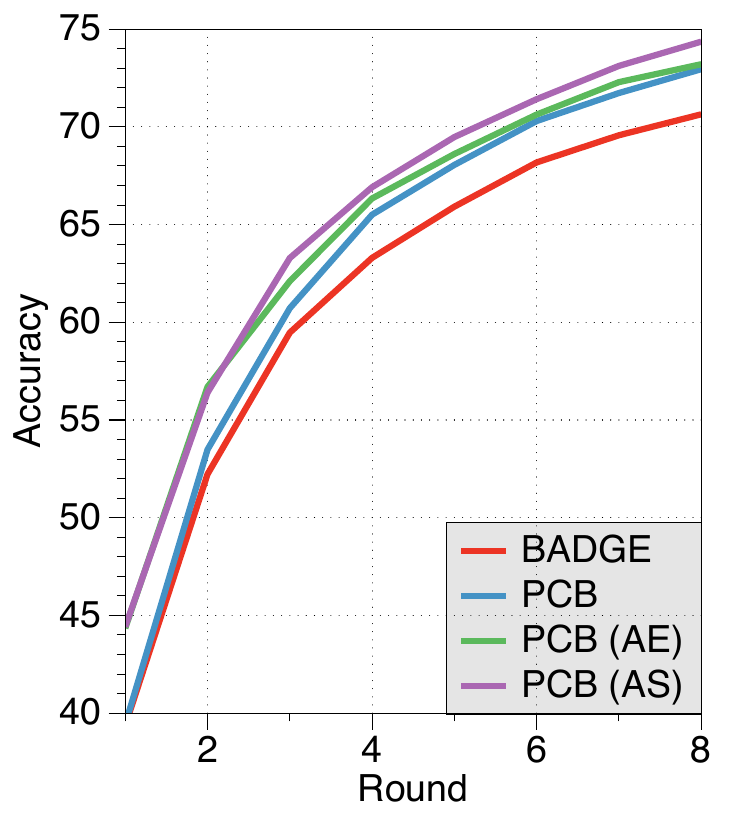}
         \vspace{-17pt}
         \caption{BADGE}
         \label{fig:badge}
     \end{subfigure}
     \vspace{-20pt}
        \caption{\textbf{Learning curve.} Average accuracy on downstream tasks with the ViT-B/32 image encoder for each round.}
     % \vspace{-3pt}
        % \caption{\textbf{Accuracy on various downstream tasks with ViT-B/32 image encoder for each round.} The graphs in terms of the other datasets are depicted in~\autoref{app:additional_results}.}
        % For the sake of space, we only show the graphs of three datasets (others in Appendix.A). 
        % As round goes on, the accuracy increases due to increasing the number of labeled dataset $\mathcal{D}_l$. }
        % For all rounds, BADGE+\alg(AS) mostly outperforms the others, but Entropy+\alg(AS) sometimes has the best performance.}
        \label{fig:round}
\end{figure}

\myparagraph{\alg improves performance.} 
We evaluate our methodology by integrating it with three active learning methods---Entropy, Coreset, and BADGE---and compare it with the Random approach and the pre-trained zero-shot CLIP model. As shown in~\autoref{tab:mainresults}, the proposed algorithm mostly improves performance in each case of its integration. For example, in the DTD dataset with the BADGE algorithm, applying \alg (\code{AS}) results in a $3.35\%$ improvement compared to the case without our algorithm. Furthermore, on average across datasets, leveraging our algorithm shows a performance improvement up to $4.64\%$. Additionally, across all active learning algorithms, \alg (\code{AS}) cases typically exhibit the highest accuracy.

\myparagraph{Active learning can be poor than Random.}
As shown in \autoref{tab:mainresults}, active learning algorithms exhibit lower performance than the Random approach in some cases. For example, especially in EuroSAT and Aircraft, both Entropy and Coreset strategies perform worse than the Random approach, because these active learning algorithms select imbalanced datasets for labeling, leading to performance degradation. See~\autoref{app:imbalance} for the detailed evidences.

\begin{figure}
     \centering
     \begin{subfigure}[b]{0.155\textwidth}
         \centering
         \includegraphics[width=\textwidth]{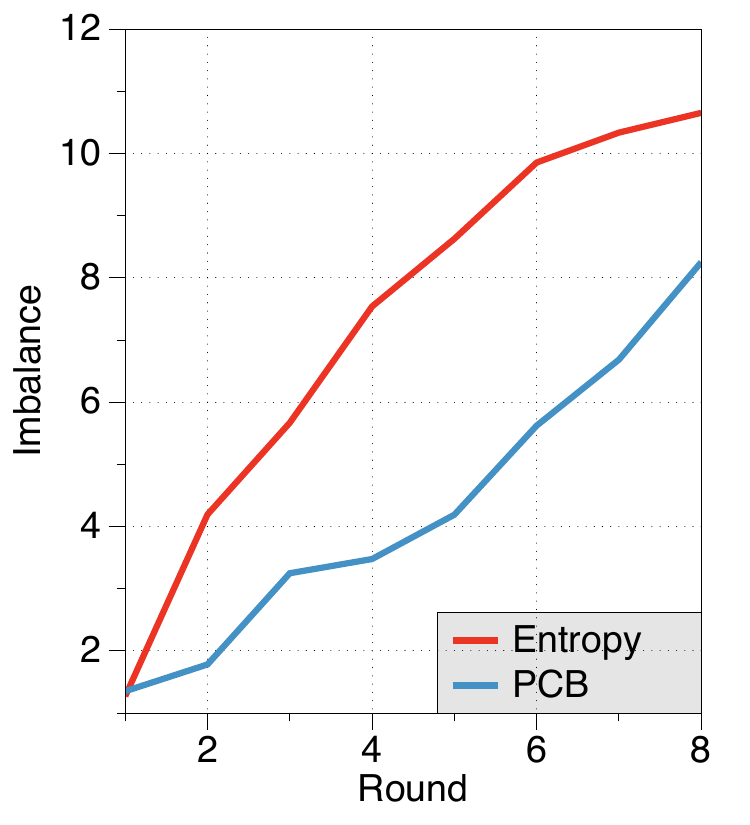}
         \vspace{-17pt}
         \caption{Entropy}
         \label{fig:entropy_var}
     \end{subfigure}
     \begin{subfigure}[b]{0.155\textwidth}
         \centering
         \includegraphics[width=\textwidth]{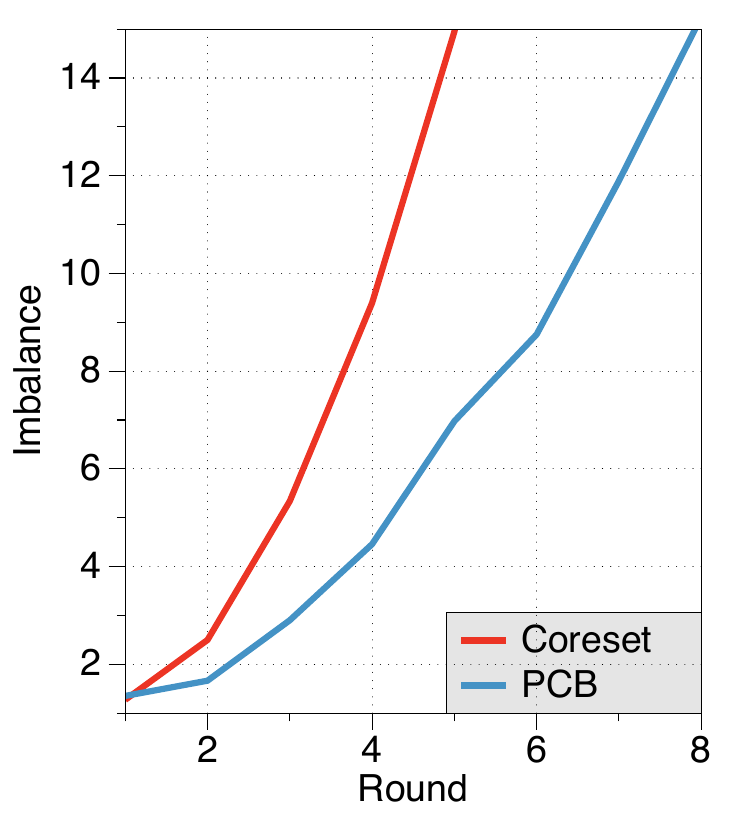}
         \vspace{-17pt}
         \caption{Coreset}
         \label{fig:coreset_var}
     \end{subfigure} 
     \begin{subfigure}[b]{0.155\textwidth}
         \centering
         \includegraphics[width=\textwidth]{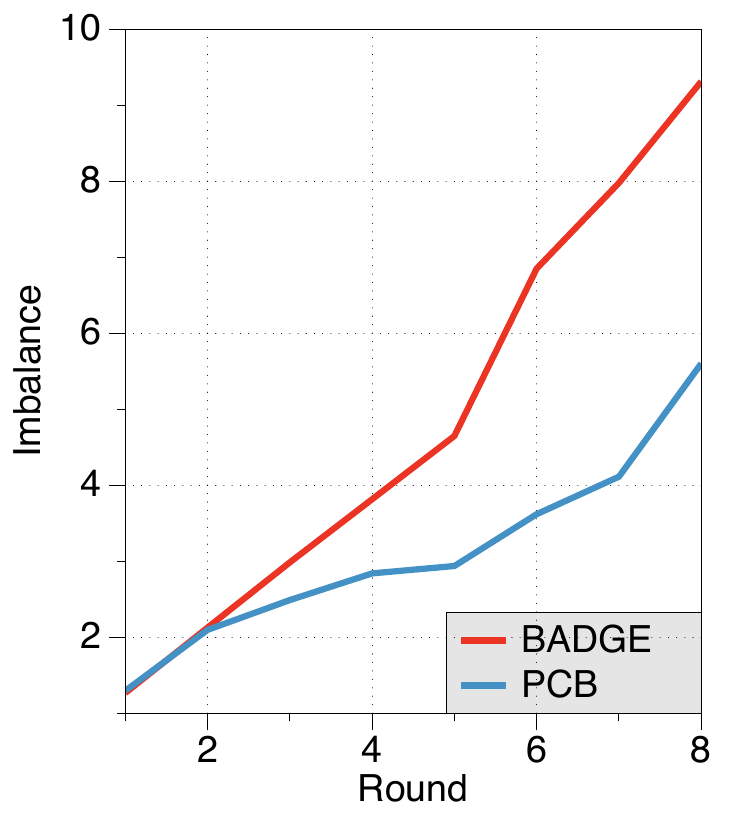}
         \vspace{-17pt}
         \caption{BADGE}
         \label{fig:badge_var}
     \end{subfigure}
     \vspace{-20pt}
        \caption{\textbf{Imbalance curve.} Average variance of the number of labeled samples for each class on downstream tasks with the ViT-B/32 image encoder for each round.}
     % \vspace{-3pt}
        % \caption{\textbf{Accuracy on various downstream tasks with ViT-B/32 image encoder for each round.} The graphs in terms of the other datasets are depicted in~\autoref{app:additional_results}.}
        % For the sake of space, we only show the graphs of three datasets (others in Appendix.A). 
        % As round goes on, the accuracy increases due to increasing the number of labeled dataset $\mathcal{D}_l$. }
        % For all rounds, BADGE+\alg(AS) mostly outperforms the others, but Entropy+\alg(AS) sometimes has the best performance.}
        \label{fig:round_imb}
\end{figure}

\myparagraph{Learning curve.} 
\autoref{fig:round} illustrates the learning curve of the avearge accuracy among various datasets for each algorithm with or without the proposed algorithm. In all cases, applying \alg with \code{AS} shows the best performance. Furthermore, utilizing \alg improves the performance compared to the active learning algorithms without \alg. In particular, all figures represent that applying \alg demonstrates an increasing gap between with \alg and without \alg as the training progresses. It indicates that, based on our imbalance analysis described in~\autoref{fig:round_imb} and detailed in~\autoref{app:imbalance}, reducing the imbalance when constructing the query set $\mc{Q}$ is crucial in active prompt learning.

% This indicates that balancing boosts up the performance due to the balance quality of labeled samples.
% This indicates that an imbalance endows the model with biased knowledge, leading to its suboptimal training. This undertrained model can exacerbate the imbalance, thereby causing a negative feedback loop. 
% This result indicates that imbalance causes misleading information imbibed by the model {\color{red}and} can trigger more severe imbalance, generating a negative spiral.
% \autoref{fig:round} illustrates the accuracy of random sampling and active learning techniques combined with our algorithm in each round. This reveals that BADGE, when mixed with our approach, consistently surpasses the rest throughout all rounds, excluding the first round. Further, upon more easily learnable dataset, Entropy+\alg(AS) reach the performance of BADGE +\alg(AS) (\autoref{fig:flower_round}).

\myparagraph{Additional analysis.}
In the case of Oxford Pets, \alg exhibits lower performance than CLIP\,(zero-shot). This result is consistent with the results from the original CoOp paper~\cite{zhou2022learning}. To further analyze this phenomenon, we increase the number of samples (\ie $N$) selected at each round by the active learning algorithm from $K$ to $16K$. When $N$=$4K$, \alg combined with BADGE outperforms zero-shot CLIP, and detailed results are described in~\autoref{app:oxford_pets}. We can conclude that the reason of performance degradation in Oxford Pets is due to the lack of samples involved in training.

\subsection{Detailed Analyses}
\label{sec:analysis}

In this section, we answer the following questions: (1) other architectures of the image encoder in the CLIP family, (2) class imbalance analysis over different $\gamma$ values, (3) analysis on various prompt learning methods, and (4) hyperparameter sensitivity analysis.

% \subsubsection{Various Image Encoder Architectures}
\myparagraph{Other types of image encoder.}
We assess the efficacy of our method across various image encoder models, as described in~\autoref{tab:various_arch}. Given the superior performance of \alg coupled with BADGE, as evidenced in~\autoref{tab:mainresults}, our subsequent analysis in~\autoref{tab:various_arch} is confined to this particular setup. Our method shows a similar trend across different encoder architectures, as supported by these observations: (1) the accuracy of zero-shot CLIP can be significantly enhanced with finetuning randomly sampled data, (2) the combination of BADGE with \alg yields better results than random sampling, and (3) description augmentation (\ie \code{AS} and \code{AE}) noticeably enhances the accuracy. 
% We note that the overall accuracies increase regardless of the method adopted, with the gap between \code{AS} and \code{AE} shrinking with an increase in the model size.
Regardless of the encoder used, the overall accuracy increases, with the difference between \code{AS} and \code{AE} decreasing as the model size grows.

\begin{figure}[t!] 
     \centering
    % \begin{subfigure}[b]{0.235\textwidth}
    %      \centering
    %      \includegraphics[width=\textwidth]{figures/flower_gamma_acc.pdf}
    %  \end{subfigure}
    %  \hfill
    %  \begin{subfigure}[b]{0.235\textwidth}
    %      \centering
    %      \includegraphics[width=\textwidth]{figures/flowers_gamma_imbal.pdf}
         
    %  \end{subfigure}
    %  \begin{subfigure}[b]{0.235\textwidth}
    %      \centering
    %      \includegraphics[width=\textwidth]{figures/dtd_gamma_acc.pdf}
         
    %  \end{subfigure}
    %  \hfill
    %  \begin{subfigure}[b]{0.235\textwidth}
    %      \centering
    %      \includegraphics[width=\textwidth]{figures/dtd_gamma_imbal.pdf}

    %  \end{subfigure}
    \centering
    \includegraphics[width=0.5\textwidth]{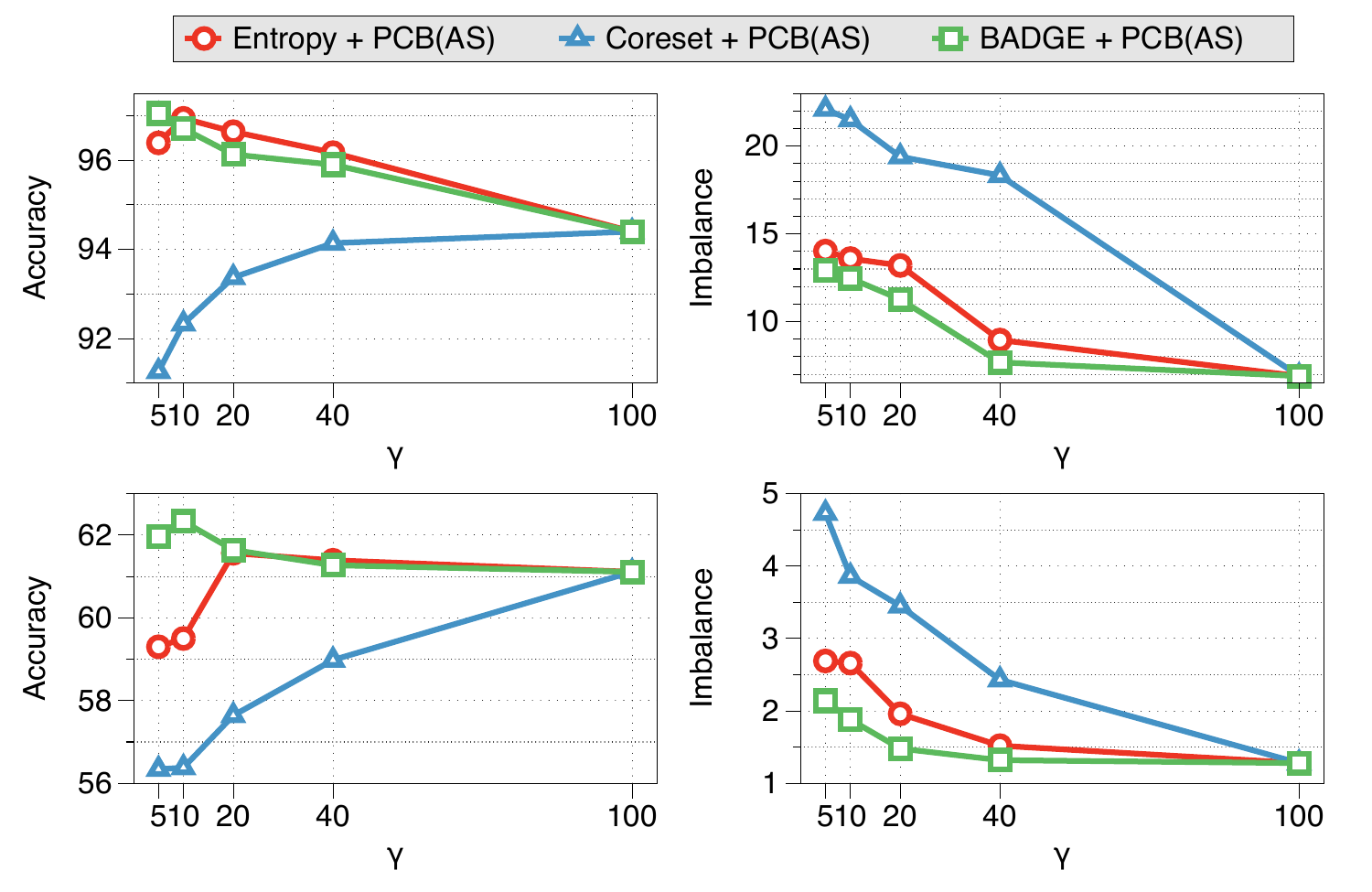}
     \vspace{-23pt}
     \caption{\textbf{Accuracy and imbalance in terms of various $\gamma$ on Flowers102 (Upper) and DTD (Bottom).}}
     % It shows that class imbalance substantially causes performance degradation, but the performance can drop when the class imbalance is lower than the certain value.}
     % \vspace{-0.5cm}
     \label{fig:gamma}
\end{figure}

% \subsubsection{Importance of Class Balance}
% \label{sec:ablation}
\myparagraph{Class imbalance analysis over different $\gamma$. }
We also study the effectiveness of our method as $\gamma$ increases and summarize the results in terms of accuracy and imbalance in~\autoref{fig:gamma}. As shown in the figure, the accuracy increases as the imbalance decreases. In particular, Coreset+\alg(\code{AS}) has higher accuracy and lower imbalance as the $\gamma$ increases due to a larger number of unlabeled examples to balance classes. On the contrary, it is noteworthy that the accuracies of Entropy+\alg(\code{AS}) and BADGE+\alg(\code{AS}) do not tend to improve as $\gamma$ increases despite of more unlabeled samples for balancing. It indicates that getting informative (\ie uncertain) data is very important to improve the accuracy after achieving a certain level of balance.

Moreover, we compare the accuracies and imbalances on two different datasets: Flowers102, which lacks class balance, and DTD, which exhibits class balance. As shown in~\autoref{fig:gamma}, the value of imbalance in Flowers102 is larger than that in DTD. More interestingly, in Flowers102, the imbalance value drops most dramatically in the range of 20\%--40\% of $\gamma$, whereas in DTD, the imbalance value drops most dramatically in the range of 10\%--20\% of $\gamma$. This indicates that achieving a class balance is obviously harder in an imbalanced dataset than in a balanced dataset.

\begin{figure}[t]
     \centering

     \begin{subfigure}[b]{0.155\textwidth}
         \centering
         \includegraphics[width=\textwidth]{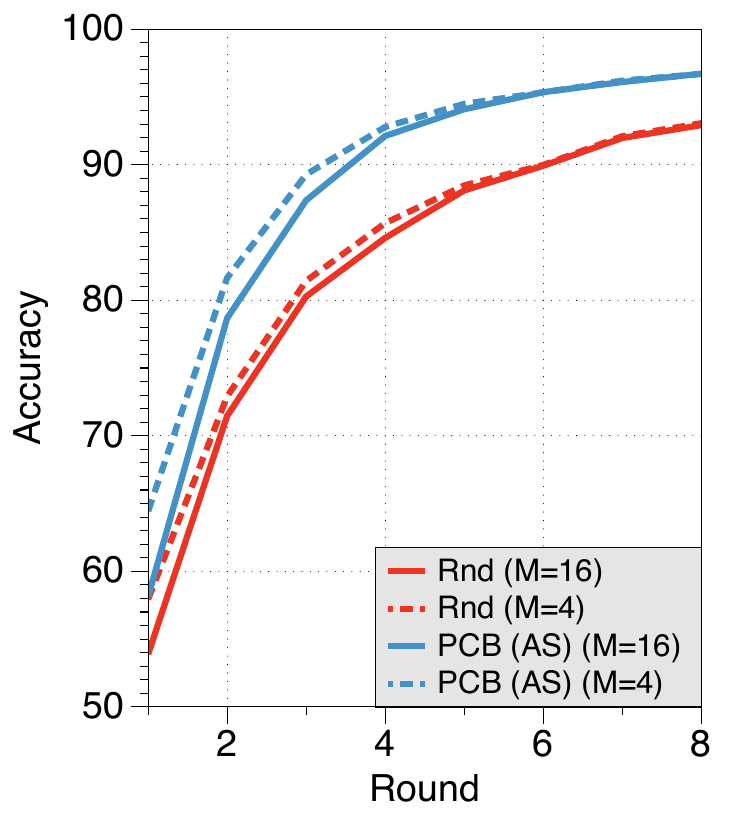}
         \vspace{-17pt}
         \caption{Prompt Size $M$}
         \label{fig:m}
     \end{subfigure}
     \hfill
     \begin{subfigure}[b]{0.155\textwidth}
         \centering
         \includegraphics[width=\textwidth]{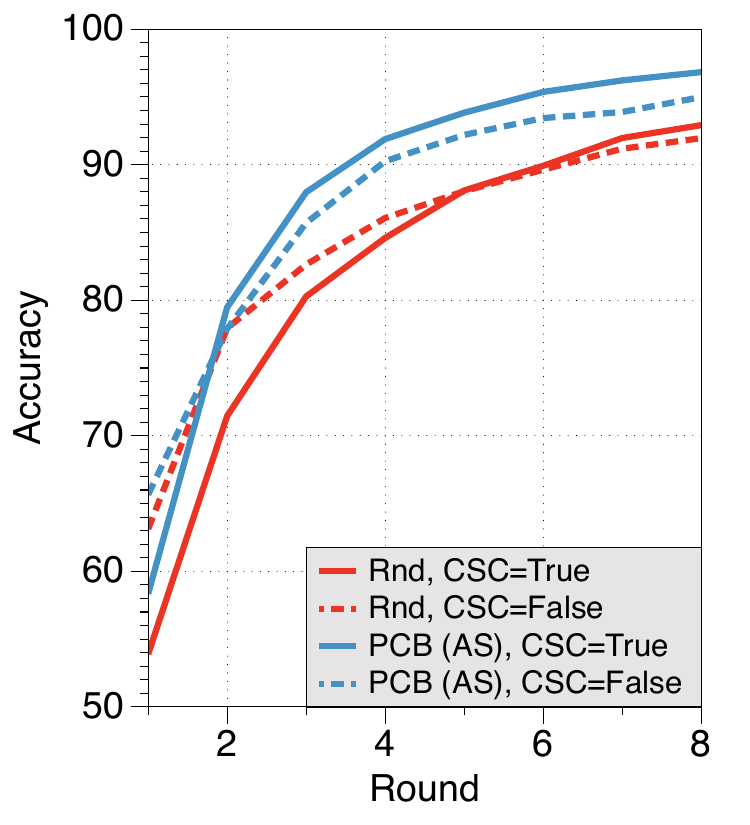}
         \vspace{-17pt}
         \caption{CSC}
         \label{fig:csc}
     \end{subfigure}
     \hfill
     \begin{subfigure}[b]{0.155\textwidth}
         \centering
         \includegraphics[width=\textwidth]{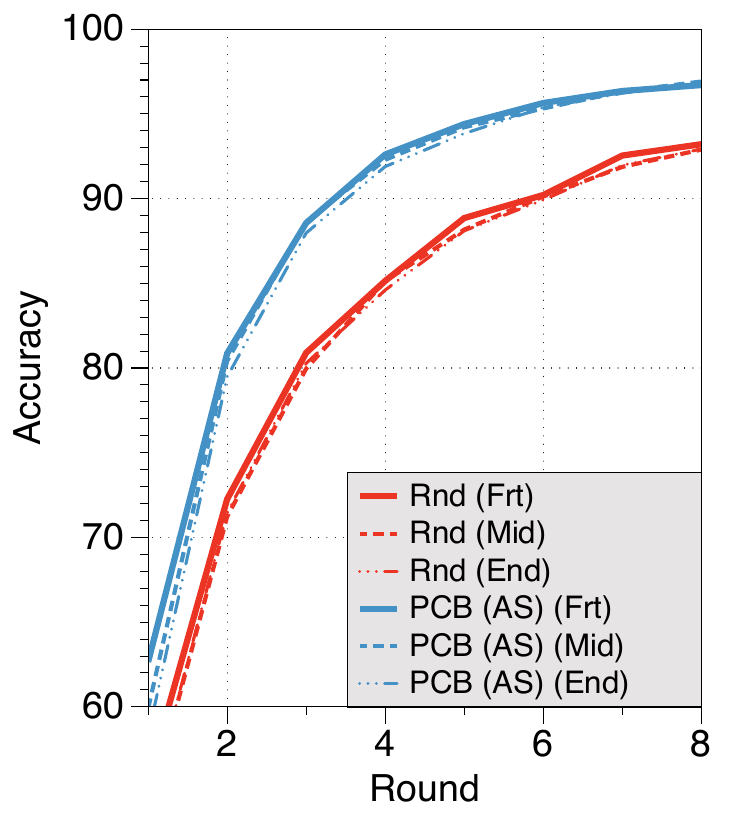}
         \vspace{-17pt}
         \caption{Prompt Position}
         \label{fig:prompt}
     \end{subfigure}
     \vspace{-20pt}
    \caption{\textbf{CoOp case analysis of BADGE on Flowers102.}}
    % :} The accuracy differences as (a) M varies,  (b) whether CSC is on or off, and (c) position of learnable context is front, middle, or end.}
     \label{fig:coop}
     % \vspace{-0.cm}
\end{figure}

% \subsubsection{Hyperparameter Studies in Prompt Learning}
\myparagraph{Hyperparameter sensitivity. }
% In this paper, we follow CoOp~\cite{zhou2022learning} training method, which is one of the popular prompt learning in VLM. We study the effect of each training hyperparameter in active learning, and summarize the results in~\autoref{fig:coop}.
We examine various variants of CoOp training methods, including different prompt sizes ($M$), cases where class-wise different tokens are not allowed (CSC=False), and variants in the position of trainable parameters (Front, Middle, and End). 

In~\autoref{fig:m}, it is observed that the accuracy with a small $M$ is higher than that with a large $M$, but this gap decreases as the rounds progress. Since the number of trainable parameters with a large $M$ is greater than that with a small $M$, the model with a large $M$ can be easily overfitted by a small labeled dataset at the initial round.

% We also analyze the performance gap between instances when the context vectors are shared for all classes (\ie CSC=False) and when different context vectors are used for each class (\ie CSC=True). We have summarized the results in ~\autoref{fig:csc}. Similar to ~\autoref{fig:m}, the accuracy when CSC is false is higher than the accuracy when CSC is true. However, as the round progress, the accuracy when CSC is true outperforms the accuracy when CSC is false. We suspect that this phenomenon originates from the difference in the number of trainable parameters, which is similar to ~\autoref{fig:m}.
We analyze the performance gap when context vectors are shared for all classes (\ie CSC=False) versus when different context vectors are used per class (\ie CSC=True), and report the results in~\autoref{fig:csc}. The accuracy is shown to be initially higher when CSC=False, but it is beaten by CSC=True as the rounds progress. This phenomenon is attributed to the difference in the number of trainable parameters simlarly to~\autoref{fig:m}.

Last, we measure the accuracy by changing the position of context vectors: Front (Frt), Middle (Mid), or End. As shown in~\autoref{fig:prompt}, the accuracies with the Front position of context vectors are slightly better than those with the others over all rounds, but the gap is within the standard deviation. As such, it is hard to conclude that the position of context vectors affects the performance in active learning.
% the position of context vectors do not affect the performance in active learning. 

\begin{table}[t]
  \centering
  \begin{adjustbox}{width=0.98\linewidth}
  \begin{tabular}{@{}cllccccc}
    \toprule
    Visual & Text & Method & $N$ & \alg & Entropy & Coreset & BADGE \\ \midrule
    \multirow{7}{*}{\rotatebox[origin=c]{90}{RN50}}   & \multirow{3}{*}{None} 
                                & LP & 102 & \xmark & 79.78\small{$\pm$1.01} & 70.66\small{$\pm$1.16} & 81.23\small{$\pm$0.40}   \\
                                & & FFT & 102 & \xmark & 47.67\small{$\pm$1.08} & 48.19\small{$\pm$2.35} & 53.43\small{$\pm$0.61} \\ 
                                & & FFT & 250 & \xmark & 77.48\small{$\pm$3.45} & 78.91\small{$\pm$0.77} & 79.51\small{$\pm$0.32} \\ \cmidrule(lr){2-8}
                                & \multirow{2}{*}{Transformer} & CoCoOp~\cite{zhou2022conditional} & 102 & \xmark & 74.54\small{$\pm$1.28} & 71.58\small{$\pm$0.73} & 78.60\small{$\pm$0.52}\\
                                & & CoOp~\cite{zhou2022learning} & 102 & \xmark & 94.74\small{$\pm$0.40} & 85.61\small{$\pm$1.36} & 95.56\small{$\pm$0.54} \\ \cmidrule(lr){2-8}
                                & \multirow{2}{*}{Transformer} & CoCoOp~\cite{zhou2022conditional} & 102 & \omark & 76.18\small{$\pm$1.55} & 72.74\small{$\pm$1.29} & 80.06\small{$\pm$1.53}\\
                                & & CoOp~\cite{zhou2022learning} & 102 & \omark & 95.89\small{$\pm$0.32} & 91.34\small{$\pm$1.00} & 95.66\small{$\pm$0.28} \\ \cmidrule(lr){1-8}
    \multirow{9}{*}{\rotatebox[origin=c]{90}{ViT-B/32}}   & \multirow{3}{*}{None} & LP & 102 & \xmark & 94.19\small{$\pm$0.77} & 86.76\small{$\pm$0.55} & 95.57\small{$\pm$0.15}\\
                                & & FFT & 102 & \xmark &37.01\small{$\pm$1.69} & 35.90\small{$\pm$1.26} & 43.14\small{$\pm$0.89} \\ 
                                & & FFT & 250 & \xmark &58.21\small{$\pm$2.89} & 58.63\small{$\pm$2.87} & 60.10\small{$\pm$0.47} \\ \cmidrule(lr){2-8}
                                & \multirow{3}{*}{Transformer} & CoCoOp~\cite{zhou2022conditional} & 102 & \xmark &76.41\small{$\pm$1.29} & 73.54\small{$\pm$0.68} & 78.94\small{$\pm$0.36}\\
                                & & MaPLe~\cite{khattak2023maple} & 102 & \xmark &84.72\small{$\pm$2.56} & 80.98\small{$\pm$0.80} & 87.86\small{$\pm$1.84}\\
                                & & CoOp~\cite{zhou2022learning} & 102 & \xmark &94.80\small{$\pm$0.75} & 88.65\small{$\pm$0.68} & 96.33\small{$\pm$0.39}\\ \cmidrule(lr){2-8}
                                & \multirow{3}{*}{Transformer} & CoCoOp~\cite{zhou2022conditional} & 102 & \omark &77.28\small{$\pm$1.71} & 73.91\small{$\pm$0.97} & 80.39\small{$\pm$0.48}\\
                                & & MaPLe~\cite{khattak2023maple} & 102 & \omark &87.60\small{$\pm$1.93} & 82.51\small{$\pm$0.22} & 88.14\small{$\pm$0.73}\\
                                & & CoOp~\cite{zhou2022learning} & 102 & \omark &96.16\small{$\pm$0.45} & 91.30\small{$\pm$0.90} & 96.12\small{$\pm$0.12}\\ \cmidrule(lr){1-8}
    \multirow{6}{*}{\rotatebox[origin=c]{90}{ViT-B/16}}   & \multirow{3}{*}{Transformer} & CoCoOp~\cite{zhou2022conditional} & 102 & \xmark &84.62\small{$\pm$1.95} & 78.44\small{$\pm$1.91} & 86.85\small{$\pm$1.21}\\
                                & & MaPLe~\cite{khattak2023maple} & 102 & \xmark &92.66\small{$\pm$1.20} & 85.54\small{$\pm$1.73} & 93.29\small{$\pm$0.39}\\
                                & & CoOp~\cite{zhou2022learning} & 102 & \xmark &97.32\small{$\pm$0.23} & 92.22\small{$\pm$2.03} & 97.97\small{$\pm$0.41}\\ \cmidrule(lr){2-8}
                                & \multirow{3}{*}{Transformer} & CoCoOp~\cite{zhou2022conditional} & 102 & \omark &85.61\small{$\pm$1.63} & 80.44\small{$\pm$0.56} & 87.41\small{$\pm$1.42}\\
                                & & MaPLe~\cite{khattak2023maple} & 102 & \omark &93.72\small{$\pm$0.95} & 87.58\small{$\pm$0.48}& 93.34\small{$\pm$1.02} \\ 
                                & & CoOp~\cite{zhou2022learning} & 102 & \omark &97.75\small{$\pm$0.08} & 94.79\small{$\pm$0.31} & 98.32\small{$\pm$0.21}\\

    \bottomrule
\end{tabular}
\end{adjustbox}
  \vspace{-0.3cm}
  \caption{\textbf{Results of various training methods on Flowers102.}}
  % Note that Transfer and FFT be the transfer learning and full finetuning only a image encoder, respectively. Training parameters of those methods are followed to CoOp. All the prompt learning methods are applied to \alg without description augmentation, and all the training parameters of the prompt learning methods are followed to each method.}
  % \vspace{-0.3cm}
  \vspace{-3pt}
  \label{tab:convention_al}
\end{table}

\myparagraph{Various prompt learning methods.}
There have been various types of prompt learning algorithms. Specifically, CoCoOp~\cite{zhou2022conditional} and MaPLe~\cite{khattak2023maple} are popular among recent approaches, and they mainly focus on transferring to unseen novel classes. We evaluate the Flower102 performance of \alg and active learning algorithms on these other prompt learning algorithms, even though they do not mainly target the case where all classes are visible at the training pahse. Furthermore, we examine the performance of full fine-tuning (FFT), which tunes all parameters and place a linear classifier on top of the model, and linear probing (LP), which trains the linear layer to adapt to the new task.

First of all, without considering transferability, \ie CoCoOp and MaPLe, CoOp shows better performance than LP and FFT. Also, we observe that the performance of CoCoOp and MaPLe is lower than that of CoOp. This observation aligns with the results reported in each paper, specifically concerning the base class performance, which pertains to the seen class during the training phase. Regardless of their performance superiority, when we compare the performance of \xmark and \omark indicating the setups without and with \alg, we find that \alg consistently improves performance. For instance, in the case of CoCoOp ViT-B/32, it enhances performance by $1.45\%$ point. % compared to the case without \alg. 

More precisely, we can conclude that FFT exhibits lower performance in a few-shot case. This phenomenon has also been reported in previous work~\cite{zhou2022conditional}, and we can attriute it to the few-shot training, as evidenced by the performance increase when we increase the number of samples from $102$ to $250$. However, it performs less effective than CoOp, indicating that prompt learning is superior to adaptation for new tasks in a few-shot perspective. Furthermore, \alg further enhances this improvement in an active learning setting.

\section{Related Work}
\label{sec:related}

\myparagraph{Vision language models (VLMs).}
To comprehend the visual and language representations, multiple approaches have been explored~\cite{lu2019vilbert, das2017visual, de2017modulating, qi2020imagebert, gan2020large, yu2021ernie, li2020unicoder}. In the stream of trials to understand both modalities at once, several years ago, CLIP~\cite{radford2021learning} emerged, drawing significant attention due to its remarkable zero-shot performance across various tasks. In a similar vein, ALIGN~\cite{jia2021scaling} was introduced, employing a comparable training methodology but featuring distinct architectural and training dataset characteristics. Unlike CLIP, ALBEF~\cite{li2021align} introduced multi-modal transformer operations applied to the outputs of two separate image and text encoders. BLIP~\cite{li2022blip} introduced a captioning module aimed at improving model performance by rectifying noisy captions. LiT~\cite{zhai2022lit} and BLIP-2~\cite{li2023blip} enhanced training efficiency by freezing specific encoder parameters. The authors of FILIP~\cite{yao2021filip} endeavored to enable the model to discern finer image details through a fine-grained, \ie patch-level, matching training approach. Florence~\cite{yuan2021florence}, on the other hand, sought to expand representations from various perspectives, such as image-to-video and so on. 
Lastly, LLaVA~\cite{liu2023visual} proposed the visual instruction tuning method using CLIP visual encoder, and it showed the state-of-the-art performance on several VLM tasks. 

\myparagraph{Prompt learning in VLMs.}
In the realm of natural language processing, there has been numerous works~\cite{shin2020autoprompt, jiang2020can, li2021prefix, zhong2021factual, lester2021power, gao2020making, khattak2023maple} aimed at enhancing the performance of language models through the optimization of prompts. 
% as studied in numerous works~\cite{shin2020autoprompt, jiang2020can, li2021prefix, zhong2021factual, lester2021power, gao2020making, khattak2023maple}. 
The primary motivation behind these works lies in the huge size of models for fine-tuning, and it is also prevalent in the VLM area. Consequently, a considerable amount of research has been dedicated to prompt learning as a means to enhance classification accuracy. CoOp~\cite{zhou2022learning} is one of representative methods and has demonstrated that a minimal number of trainable parameters suffice to adapt to a given classification task. The authors of~\cite{lu2022prompt} introduced a methodology that leverages estimated weight distributions to assign weights aimed at minimizing classification errors.

Within this framework, several studies have aimed to enhance the generalization performance in prompt learning for VLMs~\cite{zhou2022conditional, yu2023task, khattak2023maple, khattak2023self}. The primary task of these studies is to showcase a small number of classes and evaluate unseen ones. The same authors in~\cite{zhou2022learning} introduced CoCoOp~\cite{zhou2022conditional}, which incorporates a meta-network module to improve transferability. In the work presented in~\cite{yu2023task}, the authors elucidated transferability in prompt learning from a VLM perspective. Moreover, MaPLe~\cite{khattak2023maple}, a branch-aware hierarchical prompt method, was proposed where prompts for the image encoder are influenced by prompts for the text encoder. In PromptSRC~\cite{khattak2023self}, the authors highlight that previous prompt learning methods have overlooked the forgetting phenomenon during prompt training and propose an alignment-based self-regularization method to enhance transferability. 
It is important to note that this paper primarily focuses on active learning not for generalizability to new classes but for enhanced performance on the given task.
% It is noteworthy that this paper primarily focuses on active learning not for generalizability to new classes but for enhanced performance on the given task.

% In parallel directions, such as video processing, research efforts, as exemplifeid by the authors of~\cite{ju2022prompting}, have also explored similar aveneus of inquiry.

\myparagraph{Description augmentation.} Recently, generating descriptions using large language models (LLMs) has gained popularity, owing to the significant improvements in the performance of VLMs. To generate descriptions for each specific class, we asked the questions based on the specific prompt template to the LLMs. A method was proposed by \cite{menon2023visual}, where the scores obtained from different descriptions of the same class were averaged. In contrast, \cite{pratt2023does} proposed a method, called \texttt{CuPL}, that utilized the averaged embeddings from multiple descriptions for each class.
% Recently, description generation from the large language model (LLM) has become popular due to large performance enhancement of VLM. When generating description of each class, they asked the qeustions based on the specific prompt template to the LLM. \cite{menon2023visual} proposed the method that average the scores that are obtained from different descriptions of corresponding classes. On the contrary, CuPL~\cite{pratt2023does} averaged the embeddings getting from different descriptions for each class. 

\myparagraph{Active learning.} Active learning~\cite{settles2009active, ren2021survey, geifman2019deep, munjal2022towards} aims to minimize human labeling costs by identifying informative data to maximize model performance. Most of the work has generally progressed along two trajectories: (1) uncertainty-based sampling and (2) diversity-based sampling. In uncertainty-based sampling, prediction probability-based sampling methods such as soft-max confidence~\cite{lewis1994heterogeneous}, margin~\cite{roth2006margin}, and entropy~\cite{holub2008entropy} were simple yet effective. In addition, some methods performed multiple forward passes to achieve uncertainty. An intuitive approach was to receive the outputs from multiple experts~\cite{melville2004diverse, korner2006multi, beluch2018power}. Some methods~\cite{gal2017deep, houlsby2011bayesian,kirsch2019batchbald} leveraged the Monte Carlo Dropout, which obtains the stochastic results from the same model using dropout layer. On the other hand, diversity-based sampling methods~\cite{sener2018active, parvaneh2022active} were introduced using either clustering or coreset selection protocols. The coreset method~\cite{sener2018active} identified sets of examples with the greatest coverage distance across all unlabeled data. More recently, hybrid methods leveraging both uncertainty and diversity have emerged. One such method, BADGE~\cite{ash2019deep}, employed $k$-means++ clustering within the gradient embedding space.

\section{Conclusion}
\label{sec:conclusion}
% \vspace{-0.2em}
In this paper, we delve into the realm of active prompt learning within vision-language models (VLMs). Initially, we observe a misalignment between previous active learning algorithms and VLMs due to the inherent knowledge imbalance of VLMs. This imbalance consequently leads to a class imbalance of queried samples during the active learning process. To address this challenge, we introduce a novel algorithm named \alg which rectifies this imbalance by leveraging the knowledge embedded in VLMs before soliciting labels from the oracle labeler. Through extensive experiments across a range of real-world datasets, we demonstrate that our algorithm outperforms conventional active learning methods and surpasses the performance of random sampling. We believe that this framework opens up new avenues for research in the field of active learning within VLMs.

% \begin{spacing}{0}
\myparagraph{Acknowledgement.}
The third author was supported by Institute of Information \& Communications Technology Planning \& Evaluation\,(IITP) grant funded by the Korea government\,(MSIT) (No.\ 2020-0-00862, DB4DL: High-Usability and Performance In-Memory Distributed DBMS for Deep Learning and No.\ 2022-0-00157, Robust, Fair, Extensible Data-Centric Continual Learning).

% In this paper, we investigate active prompt learning in VLMs. We find out the biggest different approach of active learning in VLM compared to previous one is to consider not only uncertainty of data but also class balance of each query for performance enhancement. We also explore VLM-specific description augmentation techniques. Our method outperforms both conventional active learning techniques and random sampling in seven various datasets, and we hope that this framework optimizes the utilization of unlabeled data.

%%%%%%%%% REFERENCES

{\small
% \balance
\bibliographystyle{ieee_fullname}
\bibliography{ref}
}

\newpage
\appendix
\supptitle

This supplementray material presents additional analysis and explanation of our paper, ``\mytitle'', that are not included in the main manuscript due to the page limitation. \autoref{app:imbalance} analyses the reason why VLMs make imbalance during the active learning pipeline. \autoref{app:desc} addresses the method details to generate the descriptions of each class. \autoref{app:exp}  describes the detail experimental settings such as datsets and active learning baselines. Also, \autoref{app:large_dataset} shows the effectiveness of our method in large datasets. \autoref{app:additional_results} describes the additional results under not only BADGE active learning algorithm but also Entropy and Coreset algorithms with various architectures of an image encoder. Lastly, since \alg indicates lower performance than zero-shot pretrained CLIP, we address this phenomenon in ~\autoref{app:oxford_pets}.

\begin{table*}[t]

  \centering
  \begin{adjustbox}{width=0.95\linewidth}
  \begin{tabular}{@{}lcccccccccccccc@{}}
    \toprule
    &  \multicolumn{2}{c}{Flowers102} & \multicolumn{2}{c}{DTD} & \multicolumn{2}{c}{Oxford Pets} & \multicolumn{2}{c}{EuroSAT} &\multicolumn{2}{c}{Caltech101}  & \multicolumn{2}{c}{Stanford Cars} & \multicolumn{2}{c}{Aircraft} \\ 
    \textbf{Method}  & Acc & Imbal  & Acc & Imbal  & Acc & Imbal  & Acc & Imbal    & Acc & Imbal   & Acc & Imbal     & Acc & Imbal     \\\cmidrule(lr){1-1} \cmidrule(lr){2-3} \cmidrule(lr){4-5} \cmidrule(lr){6-7} \cmidrule(lr){8-9} \cmidrule(lr){10-11} \cmidrule(lr){12-13} \cmidrule(lr){14-15}
    CLIP (zero-shot)                    & 66.7                              & -                                 & 44.5                              & - 
                                        & 87.0                              & -                                 & 49.4                              & -
                                        & 87.9                              & -                                 & 59.4                              & - 
                                        & 21.2                              & -  \\ 
    Random                              & 92.92            & 24.31            & 58.77            & 6.77
                                        & 78.30            & 7.17             & 77.62            & 9.50
                                        & 89.55            & 48.52            & 65.96            & 8.09
                                        & 30.69            & 6.02 \\ \cmidrule(lr){1-15}
    Entropy~\cite{holub2008entropy}     & 94.80            & 20.54            & 59.18            & 6.64
                                        & 76.81            & 7.31             & 75.46            & 10.87
                                        & 91.67            & 15.03            & 66.68            & 10.69
                                        & 25.80            & 17.11  \\
    $~~ + \mathrm{\alg}$                & 96.16            & 13.41            & 59.73            & 5.62
                                        & 80.44            & 3.01             & 80.80            & 2.40 
                                        & 92.41            & 9.83             & 67.18            & 7.75 
                                        & 26.78            & 11.75  \\
    $~~ + \mathrm{\alg(\code{AE})}$     & 96.33            & 14.86            & \underline{60.07}& 6.34
                                        & 80.87            & 4.85             & \underline{81.72}& 2.53 
                                        & 93.14            & 12.16            & 66.42            & 10.13 
                                        & 27.09            & 12.38  \\ 
    $~~ + \mathrm{\alg(\code{AS})}$     & \underline{\textbf{96.94}}   & 13.59 & 59.50            & 4.94
                                        & \underline{80.94}& 2.76             & 80.75            & 3.93 
                                        & \underline{93.48}& 11.47            & \underline{68.93}          & 9.34 
                                        & \underline{27.58}& 14.23\\ \cmidrule(lr){1-15}
    Coreset~\cite{sener2018active}      & 88.65            & 30.72            & 50.39            & 39.92
                                        & 76.70            & 18.58            & 68.09            & 37.87 
                                        & 88.78            & 48.52            & 61.75            & 24.53
                                        & 24.32            & 21.58  \\
    $~~ + \mathrm{\alg}$                & 91.30            & 21.24            & 55.77            & 15.50
                                        & 76.84            & 8.74             & 77.50            & 2.07
                                        & 89.96            & 22.08            & 63.63            & 13.44
                                        & 25.38            & 14.27 \\
    $~~ + \mathrm{\alg(\code{AE})}$     & 91.70            & 21.59            & \underline{57.09}& 16.34
                                        & 78.60            & 8.97             & \underline{79.28}& 1.00 
                                        & 90.29            & 20.33            & 62.08            & 14.38
                                        & 26.19            & 14.27 \\ 
    $~~ + \mathrm{\alg(\code{AS})}$     & \underline{92.33}& 21.50            & 56.38            & 14.98
                                        & \underline{79.50}& 9.86             & \underline{79.28}& 1.13
                                        & \underline{91.70}& 22.11            & \underline{65.75}& 12.51
                                        & \underline{26.22}& 13.74 \\
    \cmidrule(lr){1-15}
    BADGE~\cite{ash2019deep}            & 96.33            & 17.89            & 58.98            & 5.90
                                        & 80.03            & 5.71             & 79.79            & 5.47
                                        & 92.54            & 13.19            & 68.07            & 5.77
                                        & 31.25            & 6.87  \\
    $~~ + \mathrm{\alg}$                & 96.12            & 13.07            & 60.28            & 5.39
                                        & 80.22            & 1.51             & \underline{\textbf{81.98}}& 1.73
                                        & 92.21            & 11.73            & 68.50            & 4.91
                                        & 31.35            & 6.46 \\
    $~~ + \mathrm{\alg(\code{AE})}$     & 96.35            & 12.69            & 61.92            & 4.56
                                        & 81.93            & 2.45             & 80.70            & 1.20
                                        & 92.52            & 12.77            & 67.70            & 4.94
                                        & 31.80            & 6.04  \\ 
    $~~ + \mathrm{\alg(\code{AS})}$     & \underline{96.71}& 12.47            & \underline{\textbf{62.33}} & 3.55
                                        & \underline{\textbf{83.16}} & 2.43   & 81.50            & 1.47
                                        & \underline{\textbf{93.85}} & 12.54  & \underline{\textbf{70.70}} & 4.52
                                        & \underline{\textbf{32.27}} & 4.98 \\ \cmidrule(lr){1-15}
    Full data                           & 97.9                              & -                                 & 74.7                               & - 
                                        & 89.3                              & -                                 & 94.5                               & -
                                        & 94.4                              & -                                 & 80.8                               & -
                                        & 43.4                              & -  \\ 
    \bottomrule
    \end{tabular}

    \end{adjustbox}
    \vspace*{-0.3cm}
  \caption{\textbf{Final accuracy and imbalance on seven downstream tasks with ViT-B/32 image encoder. }
  } 
  % \vspace*{-0.5cm}
  \label{tab:imbalance}
\end{table*}

\section{Why Imbalane Occurs in VLMs}
\label{app:imbalance}

% \begin{figure*}[t!] 
%     \centering
%     \includegraphics[width=1.0\textwidth]{figures/flower_zeroshot.pdf}
%      \vspace{-23pt}
%      \caption{Accuracy per class for pretrained CLIP (zero-shot) with ViT-B/32 image encoder on Flowers102}
%      % It shows that class imbalance substantially causes performance degradation, but the performance can drop when the class imbalance is lower than the certain value.}
%      % \vspace{-0.5cm}
%      \label{fig:zeroshot_acc}
% \end{figure*} 

\begin{figure*}[t]
     \centering

     \begin{subfigure}[b]{0.24\textwidth}
         \centering
         \includegraphics[width=\textwidth]{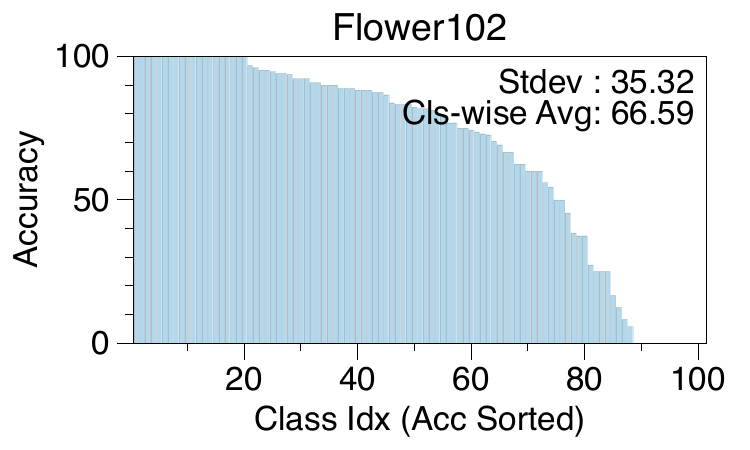}
         \vspace{-17pt}
         \caption{Flower102}
         \label{fig:percls_flower}
     \end{subfigure}
     \hfill
     \begin{subfigure}[b]{0.24\textwidth}
         \centering
         \includegraphics[width=\textwidth]{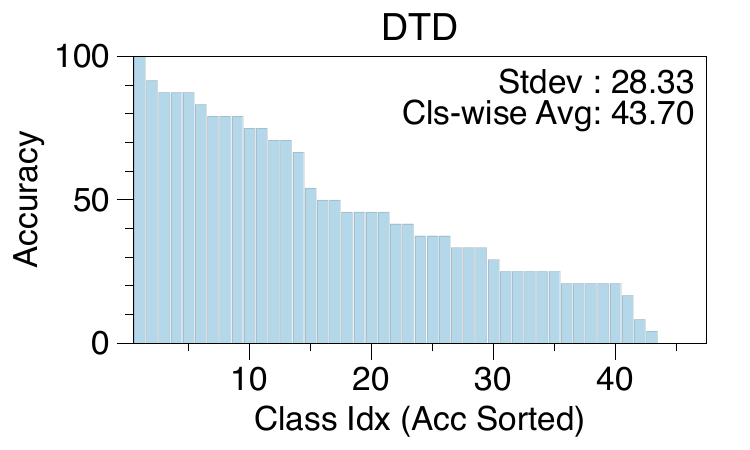}
         \vspace{-17pt}
         \caption{DTD}
         \label{fig:percls_dtd}
     \end{subfigure}
     \hfill
     \begin{subfigure}[b]{0.24\textwidth}
         \centering
         \includegraphics[width=\textwidth]{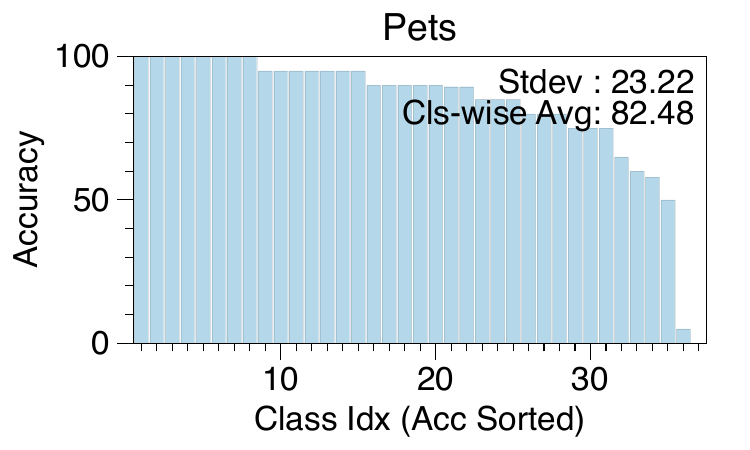}
         \vspace{-17pt}
         \caption{Pets}
         \label{fig:percls_pets}
     \end{subfigure}
      \hfill
     \begin{subfigure}[b]{0.24\textwidth}
         \centering
         \includegraphics[width=\textwidth]{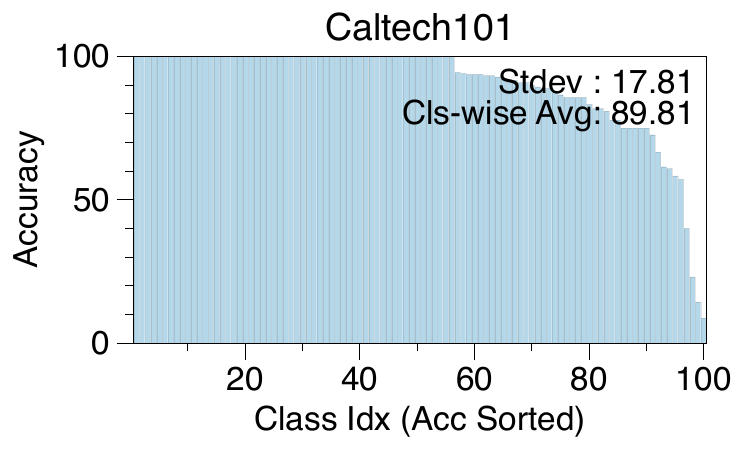}
         \vspace{-17pt}
         \caption{Caltech101}
         \label{fig:percls_caltech}
     \end{subfigure}
      \hfill
     \begin{subfigure}[b]{0.24\textwidth}
         \centering
         \includegraphics[width=\textwidth]{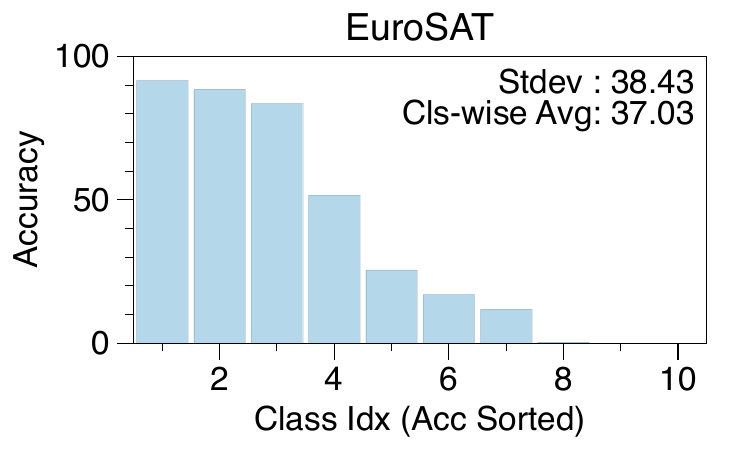}
         \vspace{-17pt}
         \caption{EuroSAT}
         \label{fig:percls_eurosat}
     \end{subfigure}
     \begin{subfigure}[b]{0.24\textwidth}
         \centering
         \includegraphics[width=\textwidth]{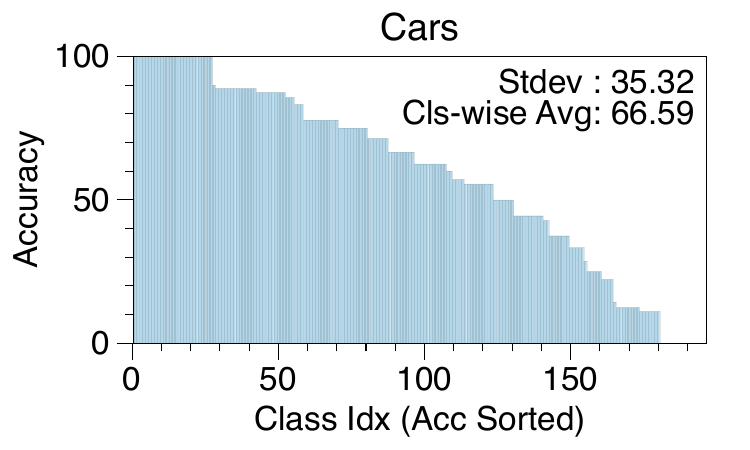}
         \vspace{-17pt}
         \caption{Cars}
         \label{fig:percls_cars}
     \end{subfigure}
     \begin{subfigure}[b]{0.24\textwidth}
         \centering
         \includegraphics[width=\textwidth]{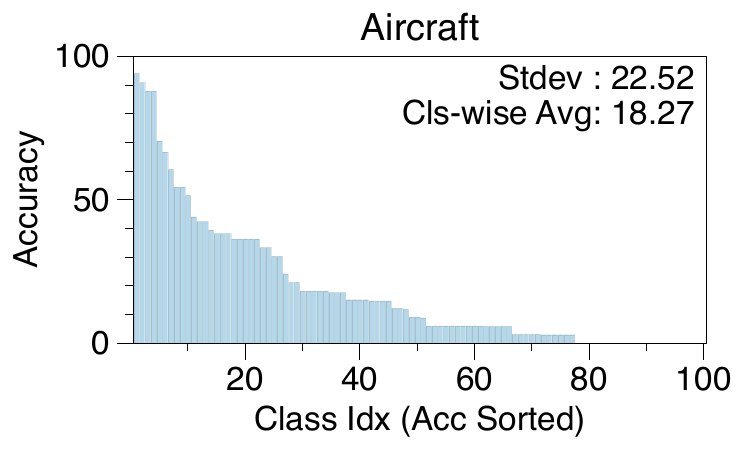}
         \vspace{-17pt}
         \caption{Aircraft}
         \label{fig:percls_aircraft}
     \end{subfigure}
     \hfill
    \caption{\textbf{Per class zero-shot accuracy from the pretrained CLIP with ViT-B/32 image encoder for each dataset.}}
     \label{fig:percls_acc}
\end{figure*}

\myparagraph{Biased knowledge of pretrained CLIP.} \autoref{fig:percls_acc} indicates the zero-shot accuracy of each class when using pretrained CLIP for all the datasets. While pretrained CLIP has powerful knowledge for some classes, it has weakness for the other classes. For instance, in the case of Flowers102, pretrained CLIP has no knowledge in terms of \textit{stemless gentian} by showing zero accuracy. On the contrary, it has perfect knowledge about \textit{moon orchid} by indicating 100\% point accuracy. As such, we can conclude that imbalanced knowledge of the pretrained CLIP causes imbalanced querying by active learning algorithms.

\myparagraph{Imbalanced dataset degrades the performance.}
\autoref{tab:imbalance} illustrates both accuracy and imbalance of labeled dataset after final round. For Oxford Pets, Stanford Cars, and FGVC aircraft datasets, where active learning algorithms can be poor than random sampling, the imbalances of Entropy and Coreset are higher than that of random sampling. It indicates that the large imbalance degrades the performance evenif they consist of uncertain or diversified data. As described in \autoref{sec:analysis}, \autoref{tab:imbalance} also shows that getting informative data enhances the accuracy after achieving a certain level of balance. For instance, despite the imbalance of BADGE being greater than that of Coreset paired with the \alg, the accuracy of the combination of Coreset and \alg is still lower than that of BADGE without \alg.

% \section{Training overhead analysis}
% \label{app:training_time}
% training overhead... 음... 이전에 괴기한 그림을 넣어야하나? 좀 더 이쁘게 그릴 순 없나...?
% 각 메소드별로 따로 그리기...? 
\section{Details for Generating Descriptions}
\label{app:desc}
\begin{figure}[t]
    \centering
    \includegraphics[width=0.9\linewidth]{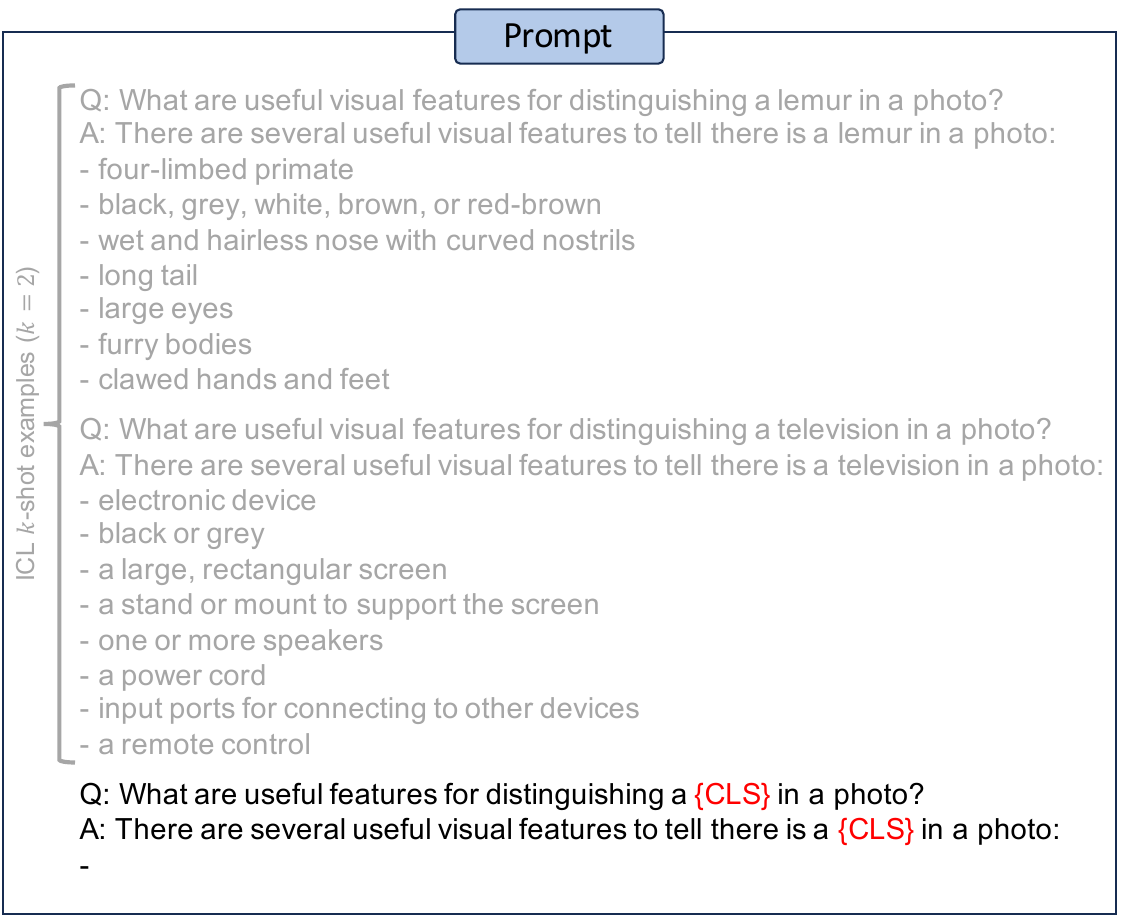}
    \caption{\textbf{Prompt template applied two-shot learning for generating descriptions.}}
    \label{fig:full_prompt}
\end{figure}

As extending \autoref{sec:desc_aug}, we delve into generating description methods in details. In NLP community, \textit{few-shot learning} is one of the popular prompt engineering skills for LLM that enhances the performance of whole tasks. It adds a few question and answer pairs, which consist of similar types of what we ask, into the prompt.
To generate the best quality descriptions for each class, we also leverage two-shot learning to LLM. Here, we show the full prompt template to get the descriptions (\autoref{fig:full_prompt}). Since the descriptions text files for DTD, EuroSAT, and Oxford Pets are included in~\cite{zhou2022learning}, we utilize them. We obtain descriptions for the remaining datasets through the use of GPT-3, whenever feasible. However, there are instances, such as with fine-grained datasets like Cars, where it proves impossible to generate descriptions for certain classes. Take, for example, the class 'Audi V8 Sedan 1994' within the Cars dataset. When prompted, GPT-3 fails to provide any description, whereas GPT-3.5-turbo produces the following output: [four-door sedan body style, Audi logo on the front grille, distinctive headlights and taillights, sleek and aerodynamic design, alloy wheels, side mirrors with integrated turn signals, V8 badge on the side or rear of the car, license plate with a specific state or country, specific color and trim options for the 1994 model year].

\section{Experimental Settings}
\label{app:exp}

\myparagraph{Datasets.} 
We select seven openly available image classification datasets that have been previous utilized in the CLIP model. Here are the details of each dataset: 

\begin{itemize}
    \item Flowers102~\cite{nilsback2008automated} consists of 102 different categoris of flowers, each representing a distinct flower species. For example, it includes roses, sunflowers, daisies, and so on. The total number of 8,189 image and label pairs. Some categories have more images than the others, which means it is imbalanced as typical real-world datasets, at least 40 and at most 258 samples.
    \item DTD~\cite{cimpoi2014describing}, abbreviated from Describable Texture Dataset, is designed for texture classification task. This dataset consists of 47 distinct classes, including categories like fabrics and natural materials. In total, DTD comprises 5,640 samples. Notably, when examining the performance reported in the CLIP~\cite{radford2021learning}, it becomes evdient that DTD poses a challenging problem for pre-trained CLIP models, as texture are not typical, easily recognizable objects.
    \item Oxford Pets~\cite{parkhi2012cats} dataset consists of 37 different pet categories, including various dogs and cats. This dataset contains 7,400 samples. Especially, it has 4,978 dog images and 2,371 cat images. We utilize class labels even though the dataset has segmentation, \ie RoI, and class both of them. 
    \item EuroSAT~\cite{helber2019eurosat} comprises 10 distinct classes that represent various land use and land cover categories. In total, this dataset includes 27,000 satelite images, with 2,700 images allocated to each of the 10 classes. Notably, each class contains an equal number of images, ensuring a balanced distribution within the dataset.
    \item Caltech101~\cite{fei2004learning} is composed of 101 unique object categories, each corresponding to a different type of object or scene. These categories encompass a wide range of objects, such as various animals, vehicles, and more. The dataset comprises a total of 9,000 images with varying numbers of images allocated to each category. Notably, it is considered a severely imbalanced dataset due to the uneven distribution of images across its categories.
    \item Stanford Cars~\cite{krause20133d} consists of a collection of 16,185 images categorized into 196 different classes, with each class typically representing a specific car make, model, and year, \eg 2012 Tesla Model S.
    \item FGVC-Aircraft~\cite{maji2013fine} encompasses a total of 10,200 images depicting various aircraft. This dataset is organized into 102 distinct classes, and each class corresponds to a specific aircraft model variant. Notably, there are 100 images available for each of these 102 different aircraft model variants. The class name in this dataset are composed of the make, model and specific variants such as Boeing 737-76J.
\end{itemize}

\myparagraph{Active learning methods.} 
To validate the effectiveness of \alg, we select three representative active learning methods: 
\begin{enumerate}
    \item Entropy~\cite{holub2008entropy} selects the most uncertain examples with the highest entropy value from logits in the prediction. Specifically, selected query set $\mc{Q}$ with size $d$ is defined as follows:
    \begin{equation*}
        \mc{Q} = \argmax_{Q \subset \mc{D}_{u}, |Q| = d} \sum_{x_i \in Q} \mc{H}\left(f(x_i)\right),
    \end{equation*}
    where $\mc{H}(f(x))$ denotes entropy of softmax output $f(x)$.
    % which is the uncertainty-based sampling method, 
    \item Coreset~\cite{sener2018active} queries the most diverse examples using embeddings from the model (\ie image encoder). More precisely, coreset is the problem that select the example that is the least relevant to the queried dataset. To solve it, the authors proposed $K$-Center-Greedy and Robust $K$-Center algorithms, and we choose the former one.
    % \smnote{조금만 더 자세히 적어주면 좋을듯!}
    % which is the diversity-based sampling method with embedding of an image encoder,
    \item BADGE~\cite{ash2019deep} considers both uncertainty and diversity by selecting the examples via $k$-means++ clustering in the gradient space. The gradient embeddings for all examples are defined as follows:
    \begin{equation}
        g_x = \frac{\partial}{\partial \theta_{\mathrm{out}}} \mc{L}_{CE} (f(x; \theta_t), \hat{y}(x))
    \end{equation}
    where $\theta_{t}$ and $\theta_{\mathrm{out}}$ refer to parameters of the model and the final layer at round $t$, respectively. $\hat{y}(x)$ denotes the pseudo label.
   
\end{enumerate}
% which is the hybrid sampling method considering both uncertainty and diversity. 
By adding \alg into those active learning methods, we study the synergy of active learning and \alg, and compare the results with random sampling (\ie instead of active learning) and zero-shot CLIP. Furthermore, we show the results when using the descriptions presented in Section~\ref{sec:desc_aug}. To illustrate the room for performance enhancement, we also measure the performance when prompt learning the model with the whole dataset (see ``Full data'').
\section{Large Dataset}
\label{app:large_dataset}

\begin{table}[t]
  \centering
  \begin{adjustbox}{width=0.95\linewidth}
  \begin{tabular}{@{}lcccc@{}}
    \toprule
    \textbf{Method}  &     ImageNet-100         &           Food101        &           SUN397         &           UCF101        \\
    \midrule
    Random           &       61.96              &            68.14         &           60.73          &           71.55          \\ 
    \cmidrule(lr){1-5}
    Entropy          &       61.26              &            66.47         &           60.57          &           71.48          \\ 
    ~~ + PCB         &       62.78              &            68.60         &           61.14          &           72.48          \\ 
    Coreset          &      61.20               &            63.76         &           55.68          &           63.47          \\ 
    ~~ + PCB         &      62.06               &            65.13         &           58.21          &           70.50          \\ 
    BADGE            &      62.66               &            69.11         &           61.84          &           74.49           \\ 
    ~~ + PCB         &      64.42               &            70.45         &           62.80          &           75.84          \\ 
    \bottomrule
    \end{tabular}
    \end{adjustbox}
    % \vspace*{-0.3cm}
    \vspace{-10pt}
    \caption{\textbf{Final accuracy with ViT-B/32 CLIP image encoder on four large scaled datasets.}}

  \label{tab:large_dataset}

\end{table}

\begin{table}[t]
  \centering
  \begin{adjustbox}{width=1.0\linewidth}
  \begin{tabular}{@{}lccccc@{}}
    \toprule
                          &     & \multicolumn{4}{c}{\textbf{Model}} \\
         \textbf{Method}  & \textbf{$N$} & RN50 & RN101 & ViT-B/32 & ViT-B/16 \\ \cmidrule(lr){1-2} \cmidrule(lr){3-6}
    CLIP (zero-shot)                    & 0          & 85.4                         &   86.2                      & 87.0                        & 88.9  \\
    Random                              & 37         & 74.65\small{$\pm$0.50}       &   79.08\small{$\pm$1.39}    & 78.30\small{$\pm$0.74}      & 84.36\small{$\pm$1.34}          \\\cmidrule(lr){1-6}
    BADGE~\cite{ash2019deep}            & 37         & 75.06\small{$\pm$0.50}       &   80.77\small{$\pm$1.31}    & 80.03\small{$\pm$1.19}      & 85.54\small{$\pm$1.30} \\
    $~~ + \mathrm{\alg}$                & 37         & 76.51\small{$\pm$1.83}       &   80.94\small{$\pm$0.42}    & 80.22\small{$\pm$1.69}      & 86.22\small{$\pm$0.71} \\ 
    $~~ + \mathrm{\alg(\code{AE})}$     & 37         & 76.77\small{$\pm$0.65}       &   83.02\small{$\pm$0.89}    & 81.93\small{$\pm$0.88}      & 87.23\small{$\pm$0.35} \\               
    $~~ + \mathrm{\alg(\code{AS})}$     & 37         & 80.09\small{$\pm$0.85}       &   83.48\small{$\pm$2.13}    & 83.16\small{$\pm$0.18}      & 88.10\small{$\pm$1.49} \\ \cmidrule(lr){1-6}
    BADGE~\cite{ash2019deep}            & 148        & 86.48\small{$\pm$0.12}       & 89.10\small{$\pm$0.16}      & 88.28\small{$\pm$0.33}      & 91.74\small{$\pm$0.23}\\
    $~~ + \mathrm{\alg}$                & 148        & 86.70\small{$\pm$0.07}       & 89.19\small{$\pm$0.21}      & 88.03\small{$\pm$0.33}      & 91.92\small{$\pm$0.81}\\ 
    $~~ + \mathrm{\alg(\code{AE})}$     & 148        & 86.54\small{$\pm$0.55}       & 89.52\small{$\pm$0.17}      & 88.30\small{$\pm$0.34}      & 91.72\small{$\pm$0.07}\\
    $~~ + \mathrm{\alg(\code{AS})}$     & 148        & 87.74\small{$\pm$0.32}       & 90.16\small{$\pm$0.12}      & 89.29\small{$\pm$0.15}      & 92.64\small{$\pm$0.14}\\ \cmidrule(lr){1-6}    
    Full data                           & -          & 88.0                         &  91.1                       &   89.3                      & 92.7 \\                                  
                                    
    \bottomrule
    \end{tabular}
    \end{adjustbox}
    % \vspace*{-0.3cm}
    \vspace*{-5pt}
    \caption{\textbf{Ablation study as increasing query size $N$ on Oxford Pets.} It shows that the small amount of training set is the crucial reason why finetuning methods underperforms zero-shot CLIP.}
  \label{tab:oxford_pets}
\end{table}

\textbf{Datasets.} We select additional four large datasets that have been previous utilized in the CLIP model. Due to limited resources, we utilized the subset that consists of 16 samples per class from the dataset. Here are the details of each dataset. 

\begin{itemize}
    \item{ImageNet-100 is a subset from ImageNet~\cite{deng2009imagenet}, consisting of randomly selected 100 categories. ImageNet is comprised of a substantial collection of images, with 1,281,167 designated for training, 50,000 set aside for validation, and 100,000 for testing purposes.}
    \item{The Food-101~\cite{bossard2014food} dataset consists of 101 food categories with 750 training and 250 test images per category, making a total of 101k images. The labels for the test images have been manually cleaned, while the training set contains some noise.}
    \item{SUN397~\cite{xiao2010sun} is the database for scene regonition. It contains 397 categories and 130,519 images.}
    \item{UCF101~\cite{soomro2012ucf101} dataset is an extension of UCF50 and consists of 13,320 video clips, which are classified into 101 categories. These 101 categories can be classified into 5 types (Body motion, Human-human interactions, Human-object interactions, Playing musical instruments and Sports). The total length of these video clips is over 27 hours. All the videos are collected from YouTube and have a fixed frame rate of 25 FPS with the resolution of 320 × 240. In this work, the middle frame of each video is used as input to the image encoder.}
\end{itemize}

\noindent \textbf{Results.} \autoref{tab:large_dataset} presents further experimental results on four large datasets, following the outcomes shown in~\autoref{tab:mainresults}. Due to limited resources, we conducted our experiments without description augmentation, applying only PCB, and all the experiments are conducted only once. When comparing these results to the baselines, we observed that using PCB has 1\%--2\% points performance improvement compared to only employing conventional active learning techniques, and it is a similar trend to what was seen in~\autoref{tab:mainresults}.
\section{Additional Results}
\label{app:additional_results}
\autoref{tab:various_arch} indicates the performance on various types of architectures of an image encoder under BADGE active learning. To extend it, we conduct the experiment on various types of architectures under not only BADGE but also Entropy and Coreset active learning algorithms, and summarize the results in \autoref{tab:app_various_arch}. Regardless of the architecture types of the image encoder, \alg combined with BADGE algorithm still has the best performance among the other baselines, but sometimes, \alg combined with Entropy algorithm beats combination of \alg and BADGE algorithm by a narrow margin. It indicates that subset $P$ sampled through the Entropy algorithm has many informative examples similar to $P$ sampled through the BADGE algorithm, where the size of $P$ is 10\% of the whole dataset.

\begin{table*}[t]
  \centering
  \begin{adjustbox}{width=0.95\linewidth}
  \begin{tabular}{@{}clcccccccc@{}}
    \toprule
    &  & \multicolumn{7}{c}{\textbf{Final Accuracy ($\uparrow$)}} &  \\
         \textbf{Model}  &\textbf{Method}  & \multicolumn{1}{c}{Flowers102} & \multicolumn{1}{c}{DTD }& \multicolumn{1}{c}{Oxford Pets}& \multicolumn{1}{c}{EuroSAT} & \multicolumn{1}{c}{Caltech101}  & \multicolumn{1}{c}{Stanford Cars} & \multicolumn{1}{c}{Aircraft} & \textbf{Avg Acc ($\uparrow$)}\\ \cmidrule(lr){1-1} \cmidrule(lr){2-2} \cmidrule(lr){3-9} \cmidrule(lr){10-10}
    \multirow{15}{*}{RN50}              & CLIP (zero-shot)                
                                       & 65.9                             & 41.7                            & 85.4                            & 41.1                            
                                       & 82.1                             & 55.8                            & 19.3                            & 55.9 \\
    & Random                           & 92.06\small{$\pm$0.54}	         & 56.62\small{$\pm$0.97}           & 74.65\small{$\pm$0.50}          &	79.10\small{$\pm$2.31}          
                                       & 84.11\small{$\pm$0.75}	         & 61.34\small{$\pm$0.57}           & 29.15\small{$\pm$0.32}          & 68.18  \\ \cmidrule(lr){2-10}
    & Entropy~\cite{holub2008entropy}  & 95.19\small{$\pm$0.09}	         & 57.62\small{$\pm$2.13}           & 72.74\small{$\pm$0.97}          &	75.73\small{$\pm$4.28}        
                                       & 88.21\small{$\pm$0.42}	         & 61.32\small{$\pm$0.80}           & 25.13\small{$\pm$0.96}          & 67.99 \\
    & $~~ + \mathrm{\alg}$             & 95.30\small{$\pm$0.59}	         & 56.44\small{$\pm$0.39}           & 75.49\small{$\pm$0.45}          &	81.69\small{$\pm$1.63}        
                                       & 88.78\small{$\pm$0.43}	         & 62.02\small{$\pm$0.17}           & 25.75\small{$\pm$0.35}          & 69.35 \\
    & $~~ + \mathrm{\alg(\code{AE})}$  & 95.75\small{$\pm$0.23}	         & 59.02\small{$\pm$0.59}           & 76.59\small{$\pm$0.12}          &	81.77\small{$\pm$1.51}        
                                       & 89.41\small{$\pm$0.53}	         & 61.05\small{$\pm$0.99}           & 26.44\small{$\pm$0.81}          & 70.00 \\
    & $~~ + \mathrm{\alg(\code{AS})}$  & \underline{96.17\small{$\pm$0.27}} & \underline{\textbf{59.34}\small{$\pm$1.09}}& \underline{78.59\small{$\pm$1.41}}&	\underline{\textbf{83.26}\small{$\pm$0.35}}        
                                       & \underline{90.49\small{$\pm$0.02}} & \underline{63.52\small{$\pm$0.31}}& \underline{26.46\small{$\pm$0.99}} & \underline{71.12} \\ \cmidrule(lr){2-10}
    & Coreset~\cite{sener2018active}   & 85.02\small{$\pm$1.51}	         & 48.74\small{$\pm$1.00}           & 69.87\small{$\pm$2.36}          &	70.02\small{$\pm$4.16}        
                                       & 83.34\small{$\pm$1.33}	         & 57.93\small{$\pm$0.56}           & 25.38\small{$\pm$0.62}          & 62.90  \\
    & $~~ + \mathrm{\alg}$             & 88.79\small{$\pm$0.98}	         & 51.63\small{$\pm$0.30}           & 71.75\small{$\pm$1.64}          &	77.74\small{$\pm$2.13}        
                                       & 85.54\small{$\pm$0.84}	         & 58.67\small{$\pm$0.37}           & 25.33\small{$\pm$0.63}          & 65.64 \\  
    & $~~ + \mathrm{\alg(\code{AE})}$  & 89.27\small{$\pm$1.69}	         & 51.69\small{$\pm$1.25}           & 73.70\small{$\pm$0.27}          &	77.74\small{$\pm$3.33}        
                                       & 86.69\small{$\pm$0.57}	         & 57.63\small{$\pm$0.55}           & 25.17\small{$\pm$0.37}          & 65.98 \\
    & $~~ + \mathrm{\alg(\code{AS})}$  & \underline{89.50\small{$\pm$1.39}} & \underline{53.15\small{$\pm$1.37}}& \underline{75.53\small{$\pm$1.64}} &	\underline{79.79\small{$\pm$1.06}}        
                                       & \underline{87.15\small{$\pm$1.14}} & \underline{60.61\small{$\pm$0.54}}& \underline{25.88\small{$\pm$0.10}}& \underline{67.37} \\ \cmidrule(lr){2-10}                                       
    & BADGE~\cite{ash2019deep}         & 95.56\small{$\pm$0.54}	         & 58.35\small{$\pm$1.20}           & 75.06\small{$\pm$0.50}          &	80.94\small{$\pm$0.55}          
                                       & 89.67\small{$\pm$0.30}	         & 63.96\small{$\pm$0.53}           & 28.12\small{$\pm$1.03}          & 70.24  \\
    & $~~ + \mathrm{\alg}$             & 95.66\small{$\pm$0.28}	         & 57.41\small{$\pm$0.17}           & 76.51\small{$\pm$1.83}          &	80.06\small{$\pm$0.97}          
                                       & 89.06\small{$\pm$0.21}	         & 63.18\small{$\pm$0.77}           & 29.23\small{$\pm$0.35}          & 70.16  \\
    & $~~ + \mathrm{\alg(\code{AE})}$  & 95.72\small{$\pm$0.31}	         & \underline{59.20\small{$\pm$1.25}}  & 76.77\small{$\pm$0.65}          & \underline{81.96\small{$\pm$0.60}}
                                       & 89.57\small{$\pm$0.19}          & 62.62\small{$\pm$0.26}           & 28.85\small{$\pm$1.59}          & 70.67 \\ 
    & $~~ + \mathrm{\alg(\code{AS})}$  & \underline{\textbf{96.18}\small{$\pm$0.07}} & 59.14\small{$\pm$1.08}  & \underline{\textbf{80.09}\small{$\pm$0.85}} & 81.60\small{$\pm$2.89}          
                                       & \underline{\textbf{90.76}\small{$\pm$0.34}} & \underline{\textbf{66.20}\small{$\pm$0.69}}& \underline{\textbf{29.61}\small{$\pm$0.78}} & \underline{\textbf{71.94}} \\ \cmidrule(lr){2-10}
    & Full data                        & 97.6                            & 71.6                             & 88.0                            & 93.6 
                                       & 92.8                            & 78.8                             & 42.6                            & 80.71 \\ \cmidrule(lr){1-10}
    
    \multirow{15}{*}{RN101}             & CLIP (zero-shot)                
                                       & 65.7                            & 43.9                             & 86.2                            & 33.1                            
                                       & 85.1                            & 62.3                             & 19.5                            & 56.54 \\ 
    & Random                           & 92.87\small{$\pm$0.43}	         & 58.29\small{$\pm$1.24}           & 79.08\small{$\pm$1.39}          & 77.21\small{$\pm$4.13} 
                                       & 87.55\small{$\pm$0.75}	         & 70.02\small{$\pm$0.36}           & 32.76\small{$\pm$0.29}          & 71.11 \\ \cmidrule(lr){2-10}
    & Entropy~\cite{holub2008entropy}  & 96.26\small{$\pm$0.11}	         & 57.17\small{$\pm$1.54}           & 78.63\small{$\pm$0.99}          &	74.88\small{$\pm$1.26}        
                                       & 91.02\small{$\pm$0.48}	         & 70.09\small{$\pm$0.16}           & 27.49\small{$\pm$0.69}          & 70.79 \\
    & $~~ + \mathrm{\alg}$             & 96.26\small{$\pm$0.25}	         & 58.81\small{$\pm$1.39}           & 80.14\small{$\pm$1.27}          &	79.91\small{$\pm$2.06}        
                                       & 91.62\small{$\pm$0.30}	         & 70.87\small{$\pm$0.45}           & 28.11\small{$\pm$0.38}          & 72.25 \\
    & $~~ + \mathrm{\alg(\code{AE})}$  & 96.47\small{$\pm$0.39}	         & 59.81\small{$\pm$1.34}           & 82.65\small{$\pm$0.99}          &	81.23\small{$\pm$1.26}        
                                       & 92.16\small{$\pm$0.90}	         & 70.14\small{$\pm$0.56}           & 27.96\small{$\pm$1.63}          & 72.92 \\
    & $~~ + \mathrm{\alg(\code{AS})}$  & \underline{\textbf{96.49}\small{$\pm$0.17}}& \underline{60.70\small{$\pm$1.09}}& \underline{\textbf{83.64}\small{$\pm$1.02}}&	\underline{\textbf{82.43}\small{$\pm$1.35}}
                                       & \underline{92.86\small{$\pm$0.20}}& \underline{73.62\small{$\pm$0.67}}& \underline{28.68\small{$\pm$0.83}} & \underline{74.06} \\ \cmidrule(lr){2-10}
    & Coreset~\cite{sener2018active}   & 87.90\small{$\pm$0.92}	         & 52.23\small{$\pm$1.76}           & 74.02\small{$\pm$1.81}          &	66.62\small{$\pm$0.54}        
                                       & 87.23\small{$\pm$1.18}	         & 65.83\small{$\pm$0.43}           & 26.37\small{$\pm$0.42}          & 65.74  \\
    & $~~ + \mathrm{\alg}$             & 91.08\small{$\pm$0.37}	         & 54.75\small{$\pm$2.93}           & 76.43\small{$\pm$1.61}          &	75.39\small{$\pm$1.94}        
                                       & 89.36\small{$\pm$0.28}	         & 66.97\small{$\pm$0.75}           & 27.28\small{$\pm$0.33}          & 68.75 \\  
    & $~~ + \mathrm{\alg(\code{AE})}$  & 91.61\small{$\pm$1.30}	         & 56.38\small{$\pm$1.55}           & 77.11\small{$\pm$1.86}          &	76.99\small{$\pm$0.65}        
                                       & 89.90\small{$\pm$0.06}	         & 65.38\small{$\pm$0.62}           & 27.72\small{$\pm$0.39}          & 69.30 \\
    & $~~ + \mathrm{\alg(\code{AS})}$  & \underline{91.80\small{$\pm$0.28}} & \underline{57.31\small{$\pm$2.07}}& \underline{81.14\small{$\pm$0.24}}&\underline{78.49\small{$\pm$1.99}}        
                                       & \underline{90.11\small{$\pm$0.30}}& \underline{69.11\small{$\pm$0.73}}& \underline{28.31\small{$\pm$0.78}} & \underline{70.90} \\ \cmidrule(lr){2-10}   
    & BADGE~\cite{ash2019deep}         & 96.26\small{$\pm$0.07}	         & 59.93\small{$\pm$1.25}           & 80.77\small{$\pm$1.31}          &	78.23\small{$\pm$2.22} 
                                       & 91.35\small{$\pm$0.32}	         & 71.43\small{$\pm$0.97}           & 32.56\small{$\pm$0.64}          & 72.93 \\
    & $~~ + \mathrm{\alg}$           & 95.79\small{$\pm$0.38}	         & 60.20\small{$\pm$1.89}           & 80.94\small{$\pm$0.42}          &	79.55\small{$\pm$1.37} 
                                       & 91.75\small{$\pm$0.44}	         & 71.35\small{$\pm$0.39}           & 32.62\small{$\pm$1.48}          & 73.17 \\
    & $~~ + \mathrm{\alg(\code{AE})}$  & \underline{\textbf{96.49}\small{$\pm$0.26}} & \underline{\textbf{62.59}\small{$\pm$0.84}}  & 83.02\small{$\pm$0.89}& \underline{81.50\small{$\pm$0.69}}
                                       & 92.51\small{$\pm$0.32}	         & 71.42\small{$\pm$0.77}	        & 32.76\small{$\pm$0.76}          & 74.33 \\ 
    & $~~ + \mathrm{\alg(\code{AS})}$  & 96.47\small{$\pm$0.18}	         & 62.17\small{$\pm$1.04}           & \underline{83.48\small{$\pm$2.13}} & 81.14\small{$\pm$1.57}	
                                       & \underline{\textbf{92.87}\small{$\pm$0.18}} & \underline{\textbf{74.04}\small{$\pm$0.39}}& \underline{\textbf{32.84}\small{$\pm$0.85}} & \underline{\textbf{75.43}} \\ \cmidrule(lr){2-10}
    & Full data                        & 97.8                            & 74.2                             & 91.1                            & 92.9 
                                       & 94.7                            & 83.7                             & 46.0                            & 82.91 \\ \cmidrule(lr){1-10}
    
    \multirow{15}{*}{ViT-B/16}           & CLIP (zero-shot) 
                                        & 70.4                             & 46.0                             & 88.9                               & 54.1 
                                        & 88.9                             & 65.6                             & 27.1                               & 63.0    \\ 
    & Random                            & 94.98\small{$\pm$0.06}	       & 62.63\small{$\pm$1.81}           & 84.36\small{$\pm$1.34}             & 81.14\small{$\pm$1.83} 
                                        & 90.95\small{$\pm$0.85}	       & 73.62\small{$\pm$0.30}           & 38.88\small{$\pm$0.25}             & 75.22 \\ \cmidrule(lr){2-10}
    & Entropy~\cite{holub2008entropy}  & 97.63\small{$\pm$0.42}	         & 62.49\small{$\pm$0.39}           & 82.56\small{$\pm$0.49}          &	77.93\small{$\pm$0.90}        
                                       & 93.04\small{$\pm$0.41}	         & 74.35\small{$\pm$0.59}           & 33.27\small{$\pm$0.72}          & 74.47 \\
    & $~~ + \mathrm{\alg}$             & 97.75\small{$\pm$0.08}	         & \underline{64.93\small{$\pm$1.02}} & 84.89\small{$\pm$0.59}          &	83.48\small{$\pm$1.37}        
                                       & 94.23\small{$\pm$0.23}	         & 75.68\small{$\pm$0.26}           & \underline{36.03\small{$\pm$0.43}} & 76.71 \\
    & $~~ + \mathrm{\alg(\code{AE})}$  & 98.06\small{$\pm$0.35}	         & 64.36\small{$\pm$0.47}           & 87.08\small{$\pm$0.90}          &	83.55\small{$\pm$1.95}        
                                       & 94.56\small{$\pm$0.34}	         & 75.15\small{$\pm$0.55}           & 35.60\small{$\pm$1.58}          & 76.91 \\
    & $~~ + \mathrm{\alg(\code{AS})}$  & \underline{\textbf{98.48}\small{$\pm$0.14}} & 63.81\small{$\pm$1.24}   & \underline{88.03\small{$\pm$0.60}}& \underline{\textbf{85.92}\small{$\pm$0.85}}       
                                       & \underline{94.89\small{$\pm$0.28}} & \underline{77.58\small{$\pm$0.43}}& 35.84\small{$\pm$1.71}          & \underline{77.79} \\ \cmidrule(lr){2-10}
    & Coreset~\cite{sener2018active}   & 92.12\small{$\pm$1.45}	         & 56.07\small{$\pm$0.90}           & 82.17\small{$\pm$1.82}          &	72.17\small{$\pm$2.72}        
                                       & 90.66\small{$\pm$0.45}	         & 70.12\small{$\pm$0.83}           & 33.28\small{$\pm$0.45}          & 70.94  \\
    & $~~ + \mathrm{\alg}$             & 94.79\small{$\pm$0.31}	         & 59.07\small{$\pm$0.63}           & 83.09\small{$\pm$1.19}          &	80.25\small{$\pm$3.12}        
                                       & 90.60\small{$\pm$0.80}	         & 71.27\small{$\pm$0.19}           & 34.06\small{$\pm$0.66}          & 73.30 \\  
    & $~~ + \mathrm{\alg(\code{AE})}$  & 94.94\small{$\pm$0.55}	         & 60.54\small{$\pm$0.86}           & 84.52\small{$\pm$0.23}          &	\underline{84.04\small{$\pm$2.92}}        
                                       & 92.15\small{$\pm$0.09}	         & 70.10\small{$\pm$1.03}           & 33.36\small{$\pm$0.03}          & 74.24 \\
    & $~~ + \mathrm{\alg(\code{AS})}$  & \underline{95.44\small{$\pm$0.82}} & \underline{61.98\small{$\pm$1.04}}& \underline{86.77\small{$\pm$0.69}}& 83.85\small{$\pm$2.45}
                                       & \underline{92.97\small{$\pm$0.29}} & \underline{72.96\small{$\pm$0.63}}& \underline{35.24\small{$\pm$0.49}}&\underline{75.60} \\ \cmidrule(lr){2-10}   
    & BADGE~\cite{ash2019deep}          & 97.97\small{$\pm$0.41}	       & 62.84\small{$\pm$2.17}           & 85.54\small{$\pm$1.30}             & 82.22\small{$\pm$1.94} 
                                        & 93.77\small{$\pm$0.51}	       & 76.55\small{$\pm$0.78}           & 39.64\small{$\pm$0.14}             & 76.93  \\
    & $~~ + \mathrm{\alg}$            & \underline{98.32\small{$\pm$0.21}}  & 64.89\small{$\pm$1.45}           & 86.22\small{$\pm$0.71}             & 81.53\small{$\pm$3.11}	
                                        & 93.75\small{$\pm$0.28}           & 76.36\small{$\pm$0.27}           & 40.20\small{$\pm$0.30}             & 77.32   \\
    & $~~ + \mathrm{\alg(\code{AE})}$        & 98.21\small{$\pm$0.21}	       & \underline{\textbf{65.25}\small{$\pm$1.28}} & 87.23\small{$\pm$0.35}             & \underline{84.04\small{$\pm$2.92}}
                                        & 94.51\small{$\pm$0.29}           & 75.84\small{$\pm$0.44}           & 39.93\small{$\pm$0.21}             & 77.86 \\ 
    & $~~ + \mathrm{\alg(\code{AS})}$        & 98.19\small{$\pm$0.17}	       & 64.95\small{$\pm$1.47}           & \underline{\textbf{88.10}\small{$\pm$1.49}} & 83.85\small{$\pm$2.45}	
                                        & \underline{\textbf{95.12}\small{$\pm$0.26}} & \underline{\textbf{78.19}\small{$\pm$0.48}}& \underline{\textbf{40.56}\small{$\pm$0.51}} & \underline{\textbf{78.42}} \\ \cmidrule(lr){2-10}
    & Full data                         & 99.0                             & 77.7                             & 92.7                               & 95.1 
                                        & 95.3                             & 85.3                             & 53.6                               & 85.53    \\ 
    \bottomrule
    \end{tabular}
    \end{adjustbox}
    \caption{\textbf{Various architectures of image encoder as an extension of \autoref{tab:various_arch}.} We include both all the conventional active learning algorithms and \alg combined with them in terms of various architectures of image encoder.}
  % \caption{\textbf{Ablation studies for various image encoder architectures.} The synergy of Entropy and Coreset with \alg for various image encoders is reported in Appendix.B for the space sake. \alg with description has generally the best performance for all the architectures. As the model size of the image encoder is bigger, the accuracy increases for all the methods, and gets closer to the upperbound.}

  \label{tab:app_various_arch}
\end{table*}
\section{Larger size of $N$ for Oxford Pets}
\label{app:oxford_pets}

As shown in \autoref{tab:mainresults} and \autoref{tab:various_arch}, zero-shot CLIP outperforms \alg combined with all the active learning algorithms in the case of Oxford Pets. Here, \autoref{tab:oxford_pets} shows that increasing query size $N$ enhances the performance. The performance when $N$ is 4 times of the number of classes (\ie 148) surpasses the performance when $N$ is the number of classes (\ie 37) with 4--7\% points for all the architectures of an image encoder. Moreover, \alg(AS) combined with BADGE algorithm when $N$=148 almost reaches the performance when training with all the data (Full data). Through this phenomenon, setting appropriate query size $N$ is important to achieve the performance that we expect, and it should be determined by learning difficulty of the dataset. 

\end{document}